\documentclass{article}

\usepackage[main, final]{neurips_2026}

\usepackage[utf8]{inputenc} 
\usepackage[T1]{fontenc}    
\usepackage{hyperref}       
\usepackage{url}            
\usepackage{booktabs}       
\usepackage{amsfonts}       
\usepackage{nicefrac}       
\usepackage{microtype}      
\usepackage{xcolor}         

\usepackage{multirow}
\usepackage{algorithm}
\usepackage{algorithmic}
\usepackage{amsmath,amssymb,amsthm}
\usepackage{wrapfig}
\usepackage{caption}
\usepackage{tikz}
\usetikzlibrary{arrows.meta, positioning}

\usepackage{makecell}
\usepackage{graphicx}
\usepackage{rotating}
\usepackage{adjustbox}
\usepackage[table]{xcolor}
\definecolor{secondbest}{RGB}{0,0,255}
\newcommand{\best}[1]{\textbf{#1}}
\newcommand{\second}[1]{\textbf{\textcolor{secondbest}{#1}}}
\newcommand{\pairbest}[1]{\cellcolor{green!25}#1}

\title{Plug-and-Play ADMM for Inverse Problems with Flow Matching Denoiser}

\author{%
  Dibyanshu Kumar \\
  TH Würzburg-Schweinfurt, CAIRO\\
  University of W\"{u}rzburg \\
  \texttt{dibyanshu.kumar@thws.de} \\
  \And
  Magda Gregorov\'{a} \\
  TH Würzburg-Schweinfurt, CAIRO\\
  \texttt{magda.gregorova@thws.de} \\
}

\usepackage{amsmath,amsfonts,bm}

\def\vtheta{{\bm{\theta}}}

\def\mI{{\bm{I}}}

\def\rvepsilon{\boldsymbol{\epsilon}}

\def\rvb{{\mathbf{b}}}

\def\rvu{{\mathbf{i}}}

\def\rvr{{\mathbf{r}}}

\def\rvu{{\mathbf{u}}}
\def\rvv{{\mathbf{v}}}

\def\rvx{{\mathbf{x}}}
\def\rvy{{\mathbf{y}}}
\def\rvz{{\mathbf{z}}}

\begin{document}

\maketitle

\begin{abstract}
Plug-and-Play (PnP) methods with generative priors have achieved strong performance 
in inverse problems, with recent flow matching models enabling fast and scalable 
reconstruction. 
However, a key challenge in PnP methods is maintaining consistency between data fidelity 
updates and the learned generative prior, as data-consistency steps can move iterates 
away from the flow trajectory.
We propose an ADMM-based PnP framework for flow matching priors that enforces stronger 
consistency between the measurement model and the prior through auxiliary variables 
and dual updates. 
In addition, we introduce a trajectory correction mechanism that restores consistency 
with the flow via a combination of deterministic and stochastic projections.
Our method provides a more principled integration of flow priors within PnP 
frameworks, achieving high-quality reconstructions with reduced computational cost.
\end{abstract}

\section{Introduction}
\label{intro}

A fundamental challenge in imaging and scientific computing is solving the inverse problem,
where the goal is to recover an unknown signal $\rvx \in \mathbb{R}^m$ from
measurements $\rvy \in \mathbb{R}^n$, modeled through a linear forward operator
$\mathcal{A} : \mathbb{R}^m \rightarrow \mathbb{R}^n$.
In practice, $\mathcal{A}$ is typically non-invertible or ill-conditioned and the measurements are noisy,
making the inverse problem ill-posed.

The measurement process is commonly described by the linear model
\begin{equation}
    \rvy = \mathcal{A}\rvx + \boldsymbol{\epsilon},
\end{equation}
where $\boldsymbol{\epsilon} \in \mathbb{R}^n$ represents measurement noise, 
typically assumed to follow a Gaussian distribution, i.e., 
$\boldsymbol{\epsilon} \sim \mathcal{N}(\mathbf{0}, \sigma^2 \mI)$.

The unknown signal $\rvx$ is modeled as a sample from a 
data distribution with density $p(\rvx)$, while the measurements $\rvy$ are generated according to the 
likelihood $p(\rvy \mid \rvx)$ induced by the forward model. 
The inverse problem then amounts to computing the posterior 
distribution $p(\rvx \mid \rvy)$. 

There are two primary approaches to solving this problem using unconditional 
pretrained diffusion models. 
The first performs posterior sampling by leveraging the learned score 
$\nabla_{\rvx_t} \log p(\rvx_t)$ and combining it with an approximation of the 
intractable likelihood score $\nabla_{\rvx_t} \log p(\rvy \mid \rvx_t)$ at each 
diffusion timestep, enabling iterative refinement 
\cite{chung2024diffusion,song2023pseudo}.

The second approach formulates the problem as an optimization task, where the prior $p(\rvx)$ is 
incorporated through regularization or proximal steps within an iterative algorithm. 
This class of methods includes MAP estimation 
\cite{zhu2023denoising,mardani2023variational} and projection-based frameworks
\cite{song2022solvinginverse,chung2024improving}, where updates alternate between enforcing 
data fidelity and applying the prior operator. 
In this setting, the reconstruction is 
formulated as the following minimization problem,
\begin{align}
\hat{\rvx}
&= \arg\min_{\rvx} \; \left\{ -\log p(\rvy \mid \rvx) - \log p(\rvx) \right\} \\
\hat{\rvx}
&= \arg\min_{\rvx} \; \left\{ l(\rvy; \rvx) + s(\rvx) \right\}
\label{mapequation}
\end{align}

where $l(\rvy; \rvx) = -\log p(\rvy \mid \rvx)$ enforces data consistency, 
and under Gaussian noise assumption reduces to 
$l(\rvy; \rvx) = \frac{1}{2\sigma^2} \|\rvy - \mathcal{A}\rvx\|^2$, 
while $s(\rvx) = -\log p(\rvx)$ is a regularizer that encodes prior information.

\cite{venkatakrishnan2013plug} introduced the Plug-and-Play (PnP) framework that replaces 
the regularizer with a denoiser-induced implicit prior, enabling the use of 
pretrained models without explicitly specifying the prior. 
Early PnP methods \cite{Zhang_2017,ryu2019plugandplay} used 
CNN denoisers for inverse problems, while later works incorporated stronger priors such as 
GANs \cite{bora2017compressed} and diffusion models 
\cite{zhu2023denoising,mardani2023variational,li2025decoupled} for improved 
reconstruction. 
More recently, motivated by the strong empirical performance of flow matching 
\cite{lipman2023flowmatching}, researchers have extended the PnP framework to flow
based models 
\cite{martin2025pnpflow,pourya2026flower}.
These methods provide a more efficient alternative by learning deterministic transport maps that can yield nearly 
straight trajectories \cite{liu2022flowstraight}, enabling faster inference 
and improved scalability for inverse problems.

\textbf{Challenges:} Despite their strong empirical performance, integrating generative priors within PnP 
frameworks remains challenging. 
A key difficulty arises from the interaction between the data-consistency step and the learned 
prior. 
In standard PnP algorithms, the data-consistency update (e.g., a gradient or proximal 
step with respect to $l(\rvy; \rvx)$) can move the iterate away from the data manifold or the learned 
flow trajectory, thereby degrading the effectiveness of the prior $s(\rvx)$. This mismatch is 
especially pronounced for flow matching models (or flow matching ODEs), where the generative 
process follows structured transport paths, and updates that are not aligned with the learned 
vector field can deviate from these trajectories.

\textbf{Contributions:} In this work, we study PnP methods with flow matching priors and make the 
following contributions, 

\begin{itemize}
    \item \textbf{ADMM-based Plug-and-Play formulation for flow priors.}
    We extend the PnP framework based on the Alternating Direction 
    Method of Multipliers (ADMM) \cite{boyd_distributed_2010} for inverse problems to 
    flow matching priors.
    ADMM introduces auxiliary variables and dual updates that explicitly enforce consistency 
    between the measurement model and the learned prior, leading to improved reconstructions
    (section \ref{analysisadmm} and table \ref{tab:results}).
    
    \item \textbf{Trajectory correction for flow consistency.}
    We introduce a new trajectory correction mechanism designed for flow-based priors. 
    Following the data-consistency step, we project the iterate back toward the flow 
    trajectory using a weighted combination of stochastic and deterministic projections. 
    This method helps recovery from poor initializations while guiding convergence 
    to consistent solutions (section \ref{analysisadmm} and table \ref{tab:results}).

    \item \textbf{Improved integration of flow priors for inverse problems.}
    By combining ADMM with our trajectory correction, we achieve a more robust 
    integration of flow-matching priors within PnP frameworks, leading to strong 
    reconstruction performance and significant reduction of computational cost, 
    runtime, and memory footprint compared to existing approaches
    (section \ref{computation} and table \ref{tab:time_memory_compact}).
    Our code is available at \url{https://github.com/dibyanshu100/flow_admm.git}. 
\end{itemize}

\section{Background}
\label{sec:bkg}

\subsection{Plug-and-Play Methods for Inverse Problems}

The PnP framework addresses inverse problems by combining a 
data-fidelity term with a learned prior implemented as a denoising operator. 
A key feature of PnP is that the prior is trained independently of the inverse 
problem, typically as a standalone denoiser, and can therefore be reused across 
different forward models. 
This decoupling enables the use of powerful learned 
priors without requiring problem specific retraining.

PnP methods have been developed across a range of optimization schemes. 
Early works employed CNN-based denoisers within ADMM \cite{venkatakrishnan2013plug}, 
Half-Quadratic Splitting (HQS) \cite{zhang2017learning}, 
Regularization by Denoising (RED)\cite{romano2017little}, 
and Forward Backward Splitting (FBS) \cite{ryu2019plugandplay}. 
More recently, stronger generative priors have been incorporated, with diffusion-based 
denoisers adopted within HQS \cite{zhu2023denoising}.
These methods share a common structure of alternating updates between 
data-consistency and denoising steps, but differ in how they enforce 
data fidelity and integrate the learned prior.
Notably, this distinction becomes more pronounced for complex generative priors such as diffusion or flow models, 
where consistency with the underlying data trajectory is not explicitly enforced.

In this work, we adopt a general operator-based view of PnP methods, where the 
prior is represented by a denoising operator $\mathcal{D}_t(\cdot)$. 
In particular, we consider a time-dependent denoiser derived from a pre-trained flow 
matching model, thereby integrating flow-based priors within the PnP 
framework.

\subsection{Flow Matching and Straight Trajectories}

Flow matching \cite{lipman2023flowmatching} provides a framework for learning deterministic transformations 
that transport samples from a simple base distribution $\rvx_0 \sim p_0$ 
where $p_0$ is typically a standard Gaussian, to a target data distribution 
$\rvx_1 \sim p_1$. This is achieved by learning a time-dependent 
vector field $\rvv(\rvx_t,t)$ that defines the ODE
\begin{equation}
\frac{d\rvx_t}{dt} = \rvv(\rvx_t, t), 
\quad t \in [0,1],
\end{equation}
whose solution maps $p_0$ to $p_1$.

Flow matching constructs a family of intermediate distributions 
$(p_t)_{t \in [0,1]}$ connecting $p_0$ and $p_1$. 
Given a coupling $\pi \in \Gamma(p_0, p_1)$, an effective choice of trajectory is to define 
paths via straight-line interpolation \cite{liu2022flowstraight}, i.e.\,
$\rvx_t = (1 - t)\rvx_0 + t \rvx_1$ where $(\rvx_0, \rvx_1) \sim \pi$. 
Under mild conditions, this path has a velocity field $\rvv_t$ 
satisfying the continuity equation
$\partial_t p_t + \nabla \cdot (p_t \rvv_t) = 0$.

The associated flow can be characterized by a flowmap $\psi_t(\rvx_0)$ solving
\begin{equation}
\frac{d}{dt} \psi_t(\rvx_0) = \rvv_t(\psi_t(\rvx_0)), \quad \psi_0(\rvx_0) = \rvx_0,
\end{equation}
such that $p_t = (\psi_t)_\# p_0$.
In practice, $\rvv_t$ is unknown and we approximate it using a 
neural network $\rvv_\vtheta$. 
For straight-line trajectories, the target velocity is given as
$\rvv_t^*(\rvx_t \mid \rvx_0, \rvx_1) = \rvx_1 - \rvx_0$, leading to the 
conditional flow matching objective
\begin{equation}
\mathcal{L}_{\text{CFM}}(\vtheta) = \mathbb{E}_{(\rvx_0, \rvx_1) \sim \pi, t\sim \mathcal{U}(0,1)}
\left[\left\| \rvv_\vtheta\big((1 - t)\rvx_0 + t \rvx_1, t\big) - (\rvx_1 - \rvx_0) \right\|^2\right].
\end{equation}
For inference, the flow from time $t$ to $1$ can be written as
$\psi_1(\rvx_t) = \rvx_t + \int_t^1 \rvv_\vtheta(\rvx_q, q)\, dq$.
This integral can be approximated using an explicit Euler step, 
yielding the one-step update
\begin{equation}
\psi_1(\rvx_t) \approx \rvx_t + (1 - t)\, \rvv_\vtheta(\rvx_t, t).
\label{one_step_denoiser}
\end{equation}
The resulting flow operator $\psi_1$ can be interpreted as a one-step, time-dependent denoiser, 
mapping $\rvx_t$ toward the data distribution at $t=1$. 
In our framework, we use $\psi_1$ as the denoising operator, i.e., $\mathcal{D}_t(\rvx_t) := \psi_1(\rvx_t)$, 
and describe its role within the PnP-ADMM framework in Section \ref{methods}.

\subsection{Plug-and-Play Flow via FBS}
Recently, several works have used flow-matching priors to solve inverse problems
\cite{pokle2024training,benhamu2024dflow}.
A recent approach, proposed by \cite{martin2025pnpflow}, integrates these priors into the 
PnP framework based on the Forward Backward Splitting (FBS) algorithm (see appendix \ref{fbs_algo}); we refer to it as PnP-FBS. 
This has since been adopted and further developed within various 
flow based PnP frameworks \cite{li2025imagerestorationprimaldual,wickremasinghe2025flow}.

In vanilla FBS, the optimization is performed via alternating gradient and prior 
updates in a sequential manner.  
However, as noted by \cite{martin2025pnpflow}, the data-consistency step moves the 
iterate away from the flow trajectory $p_t$. 
To address this drift, they introduce a projection 
step before applying the denoiser in PnP-FBS. 
Specifically, the output of the data-consistency step $\rvz^{k}$,
is projected back on the flow trajectory using $\rvepsilon \sim p_0$ as,
\begin{equation}
\tilde{\rvz}^{k} = (1 - t_k)\rvepsilon + t_k \rvz^{k}
\label{fbs:projection}
\end{equation}
This step should ensure that the input to the flow operator remains consistent with
the interpolation structure used during training.

\textbf{Limitations of PnP-FBS}: 
\begin{itemize}
\item \textbf{Projection inconsistency:}
The projection step in the PnP-FBS algorithm introduces stochasticity 
at every iteration through independent sampling from $p_0$. 
Since a new random sample $\rvepsilon$ is drawn at each step, the iterate is repeatedly 
perturbed away from its current trajectory.
While such stochasticity is beneficial in the early stages of the sampling process, 
where it encourages exploration across possible flow trajectories,
it becomes undesirable near convergence, where the updates tend to average out the 
solution rather than refine it toward the MAP estimate. 
As a result, the updates lack consistency across iterations, leading to deviations 
from a coherent transport path and potential loss of fine details as can be seen in the 
reconstructed images of PnP-FBS in fig. \ref{fig:results}.

\item \textbf{Step-size sensitivity and lack of consensus enforcement:}
As noted by \cite{martin2025pnpflow},
PnP-FBS is highly sensitive to the choice of the time-dependent step 
size $\gamma(t_k)$ in the gradient step.
In their work, $\gamma(t_k) = (1 - t_k)^\alpha$ with 
$\alpha \in (0,1]$, and the authors acknowledge that more effective 
tuning strategies may be required.
This sensitivity arises because, $\gamma(t_k)$ directly controls the balance 
between the data-consistency term and prior-consistency, without any explicit mechanism 
to enforce this balance. 
This makes the method rely entirely on careful step-size tuning across time, 
where an improper choice can lead to dominance of the data-consistency term and collapse 
to a noisy solution.
\end{itemize}

\section{Proposed Approach}
\label{methods}
We propose an ADMM-based plug-and-play framework with flow-matching priors, along with a structured projection  
mechanism for trajectory correction that improves stability and consistency. 
In this section, we present our approach in detail.

\subsection{Plug-and-Play Flow via ADMM}

A principled approach to incorporate denoisers into inverse problems is via ADMM, 
as proposed by \cite{venkatakrishnan2013plug}.
PnP-ADMM reformulates the MAP objective in equation \eqref{mapequation} as a constrained optimization problem by 
introducing an auxiliary variable $\rvz$:
\begin{equation}
\min_{\rvx, \rvz} \; l(\rvy;\rvx) + s(\rvz)
\quad \text{s.t.} \quad \rvx = \rvz
\end{equation}
This variable splitting decouples the data-consistency term and the prior, allowing each 
to be handled independently.
The resulting constrained problem is solved using ADMM by forming the augmented Lagrangian as,
\begin{equation}
\mathcal{L}_\rho(\rvx, \rvz, \rvu)
= l(\rvy;\rvx) + s(\rvz)
+ \frac{\rho}{2} \|\rvx - \rvz + \rvu\|_2^2,
\end{equation}
where $\rvu$ is a scaled dual variable and $\rho > 0$ is a penalty parameter that 
controls the strength of the consensus constraint.
The ADMM iterations take the form
\begin{align}
\rvx^{k+1}
&= \arg\min_{\rvx} \;
l(\rvy;\rvx)
+ \frac{\rho}{2} \|\rvx - (\rvz^k - \rvu^k)\|_2^2
\label{eq:x_update} \\
\rvz^{k+1}
&= \arg\min_{\rvz} \;
s(\rvz)
+ \frac{\rho}{2} \|(\rvx^{k+1} + \rvu^k) - \rvz\|_2^2
\label{eq:z_update} \\
\rvu^{k+1}
&= \rvu^k + \rvx^{k+1} - \rvz^{k+1}
\label{eq:u_update}
\end{align}

The $\rvz$-update corresponds to the proximal operator of the prior $s$ 
and has the same form as a Gaussian denoising problem, 
with $\rvx^{k+1} + \rvu^k$ serving as the input to the denoising operator. 
The PnP framework replaces this proximal operator in equation \eqref{eq:z_update} 
with a generic denoiser $\mathcal{D}(\cdot)$, yielding the update
\begin{equation}
\rvz^{k+1} = \mathcal{D}(\rvx^{k+1} + \rvu^k)
\label{eq:z_update_denoiser}
\end{equation}
In our approach, this denoiser is instantiated as the time-dependent flow operator 
$\mathcal{D}_t(\cdot)$ as formulated in equation \eqref{one_step_denoiser}.

For linear degradation operators $\mathcal{A}$, the data-consistency subproblem 
(equation \eqref{eq:x_update})
reduces to a quadratic objective that can be efficiently solved using iterative linear solvers, 
such as conjugate gradient \cite{hestenes1952methods}, without requiring explicit 
matrix inversion (see appendix \ref{appendix:cgd}). 
This leads to computationally efficient and scalable optimization, even in high-dimensional problems.

\paragraph{Advantages of ADMM:} The ADMM iterations can be interpreted as comprising three interacting components: 
(i) a data-consistency update defined by the forward model \eqref{eq:x_update}, 
(ii) a prior-consistency update implemented via a denoising operator \eqref{eq:z_update_denoiser}, and 
(iii) a dual variable update that enforces agreement between the two \eqref{eq:u_update}. 
The presence of the dual variable $\rvu$ is a key distinguishing feature of ADMM.
It behaves like an error signal with memory that accumulates discrepancies 
across iterations and acts as a feedback mechanism to correct future updates so that the 
two variables $\rvx$ and $\rvz$ converge to a common solution.
In contrast to FBS, which relies on a carefully tuned 
time-dependent step size $\gamma_k$, ADMM uses a penalty parameter $\rho$ that can be kept 
fixed across iterations and requires minimal tuning (see appendix~\ref{rho_analysis}).
This leads to improved empirical stability compared to FBS 
and makes ADMM-based PnP methods particularly well-suited for integrating 
complex learned priors (see evaluation in section \ref{analysisadmm}).

\paragraph{Convergence:} 
\label{conv_analysis}
Classical ADMM has convergence guarantees in convex settings
~\citep{boyd_distributed_2010}, 
but in PnP-ADMM the proximal operator is replaced by a learned denoiser, 
so standard theory no longer applies. 
Existing PnP analyses therefore establish fixed-point 
convergence under structural assumptions on the denoiser, such as nonexpansiveness or 
averagedness~\citep{chan2016plug,ryu2019plugandplay}, while stronger guarantees require 
more restrictive denoiser classes~\citep{hurault2022gradient}. 
For our flow-matching denoiser, these properties are not guaranteed, so we complement 
the theory with empirical convergence diagnostics (see appendix \ref{app:conv}), 
where the decay of primal and dual residuals indicates stable behavior in practice.

\begin{algorithm}[t]
\caption{PnP-Flow ADMM with time-adaptive projection}
\label{algo:pnpflow_admm}
\begin{algorithmic}[1]
\STATE \textbf{Input:} Denoiser $\mathcal{D}_t$, observation $\rvy$, data-fidelity $l(\rvy;\rvx)$, 
penalty parameter $\rho$, time sequence $(t_k)_k$ with $t_k = k/N$, $N \in \mathbb{N}$, or 
infinite with $\lim_{k\to\infty} t_k = 1$
\STATE \textbf{Initialize:} $\rvx^0 \sim \mathcal{N}(0,I)$, $\rvz^0=\rvx^0$, $\rvu^0=0$, $\rvepsilon^0 = \rvx^0$

\FOR{$k = 0,1,\dots$}
    \STATE $\rvx^{k+1} = \arg\min_{\rvx}\; l(\rvy;\rvx) + \frac{\rho}{2}\|\rvx-(\rvz^k-\rvu^k)\|_2^2$ \hfill $\triangleright$ data-consistency
    \STATE $\hat{\rvz}^{\,k} = \rvx^{k+1} + \rvu^k$ \hfill  $\triangleright$ consensus
    \STATE $\tilde{\rvz}^{\,k} = (1-t_k)\rvepsilon^k + t_k \hat{\rvz}^{\,k}$ \hfill $\triangleright$ projection on flow trajectory
    \STATE $\rvz^{k+1} = \mathcal{D}_{t_k}(\tilde{\rvz}^{\,k})$ \hfill $\triangleright$ denoiser prior consistency
    \STATE $\rvepsilon_P = \frac{\tilde{\rvz}^{\,k} - t_k \rvz^{k+1}}{1-t_k}, \qquad \rvepsilon_R \sim \mathcal{N}(0,I)$ \hfill $\triangleright$ deterministic\eqref{eq:deterministic} and stochastic component
    \STATE $\rvepsilon^{k+1} = \frac{t_k \rvepsilon_P + (1-t_k)\rvepsilon_R}{\sqrt{t_k^2 + (1-t_k)^2}}$ \hfill $\triangleright$ time adaptive latent sample \eqref{eq:adaptive_projection}
    \STATE $\rvu^{k+1} = \rvu^k + \rvx^{k+1} - \rvz^{k+1}$ \hfill $\triangleright$ dual update
\ENDFOR
\STATE \textbf{return} $\rvz^{k+1}$
\end{algorithmic}
\end{algorithm}

\subsection{Trajectory Correction}

While the ADMM formulation enforces consensus between the data-consistent iterate $\rvx^{k+1}$ and 
the prior-consistent variable $\rvz^{k+1}$, it does not guarantee that the denoiser input lies on the 
family of intermediate distributions $\{p_t\}_{t\in[0,1]}$ induced by the flow matching model. 
In particular, the input to the denoiser,
\begin{equation}
\hat{\rvz}^k = \rvx^{k+1} + \rvu^k,
\end{equation}
is influenced by both the forward operator and the dual variable, and therefore generally deviates 
from the linear interpolation structure used during flow training.

To address this mismatch, we introduce a \emph{time-adaptive projection} mechanism that explicitly steers 
the iterates toward the learned flow trajectory. 
Similar to the correction step in equation \eqref{fbs:projection}, 
we project $\hat{\rvz}^k$ onto the flow path via
$\tilde{\rvz}^k \leftarrow  (1 - t_k)\rvepsilon^k + t_k \hat{\rvz}^k$
where the variable $\rvepsilon^k$ represents a latent sample from $p_0$.

\paragraph{Deterministic and stochastic components of $\rvepsilon^k$:}
A key difference in our method is that we do not rely solely on random sample from $p_0$. 
Instead, we construct $\rvepsilon^k$ as a combination of:
(i) a stochastic component $\rvepsilon_R \sim \mathcal{N}(0, I)$, promoting exploration across trajectories, and  
(ii) a deterministic component $\rvepsilon_P$, that captures the local 
structure of the current flow trajectory.

To construct the deterministic component $\rvepsilon_P$, we use the coupling between $p_0$ and $p_1$ 
underlying the flow. 
Given $\tilde{\rvz}^k$ and the denoised output $\rvz^{k+1} = \mathcal{D}_{t_k}(\tilde{\rvz}^k)$, 
we estimate the corresponding latent variable in $p_0$ that is coupled with $\rvz^{k+1}$ under
the linear interpolation structure of the flow.
This is obtained from the interpolation defining the flow at time $t_k$,
\begin{equation}
\rvepsilon_P = \frac{\tilde{\rvz}^k - t_k \rvz^{k+1}}{1 - t_k}
\label{eq:deterministic}
\end{equation}

\begin{figure}[t]
    \centering
    \includegraphics[width=0.75\linewidth]{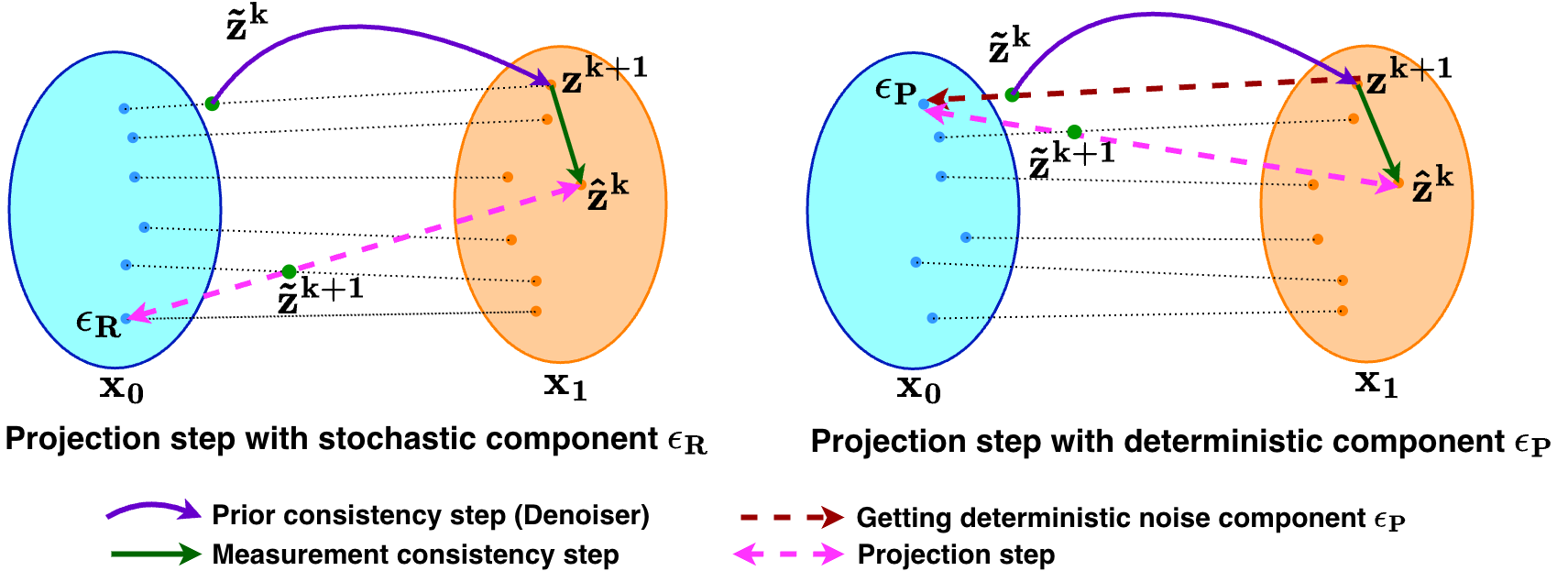}
    \caption{Comparison of projection using stochastic (left) and deterministic (right) components 
    of $\rvepsilon_k$.
    Stochastic projection using $\rvepsilon_R$ could shift the iterate across trajectories, 
    while deterministic projection using $\rvepsilon_P$ keeps it closer to the current flow path.}
    \label{fig:projection}
\end{figure}

In fig. \ref{fig:projection} we show a pictorial representation of projection using 
$\rvepsilon_P$ and $\rvepsilon_R$.
Using $\rvepsilon_P$ allows us to project the output of the data-consistency step onto a flow 
trajectory consistent with the current iterate, in contrast to the random sample $\rvepsilon_R$, 
which ignores this structure.
However, relying exclusively on $\rvepsilon_P$ can be limiting as it biases the algorithm 
toward a single trajectory determined by earlier iterates, 
potentially preventing exploration of alternative solutions that better satisfy the 
measurement constraints. 
On the contrary, using purely random sample $\rvepsilon_R$ disregards the current trajectory and may lead 
to unstable updates.
We further analyze these design choices in section \ref{analysisadmm}.

\paragraph{Time-adaptive projection:}
To balance exploration and consistency, we combine the stochastic and deterministic
component through a normalized, time-dependent weighting as,
\begin{equation}
\rvepsilon^{k+1} = \frac{t_k \rvepsilon_P + (1 - t_k)\rvepsilon_R}{\sqrt{t_k^2 + (1 - t_k)^2}},
\quad \text{with } \rvepsilon^0 = \rvx^0
\label{eq:adaptive_projection}
\end{equation}
This design follows a simple principle: at early iterations (small $t_k$), the algorithm emphasizes 
stochastic exploration via $\rvepsilon_R$, allowing the iterate to traverse different regions of 
the solution space.
As $t_k \to 1$, the weight progressively shifts toward $\rvepsilon_P$, encouraging convergence along a 
consistent flow trajectory aligned with the denoiser.
Overall, this method enables the algorithm to both 
(i) recover from poor initializations and 
(ii) commit to a stable trajectory as it approaches a solution that satisfies the data fidelity term.

\section{Experiments and Results}
\label{experiments}

\paragraph{Dataset and evaluation metrics:} 
\label{data}
We conduct our experiments on two datasets CelebA \cite{yang2015facialparts} (resized to $128 \times 128$)
and AFHQ-Cat (resized to $256 \times 256$) \cite{choi2020starganv}.
For quantitative evaluation, we randomly select $100$ test images from each dataset.
All methods are evaluated using pre-trained flow matching models from \cite{martin2025pnpflow}.
To assess reconstruction quality, we report both pixel-level and perceptual metrics. 
Specifically, we use Peak Signal-to-Noise Ratio (PSNR) and Structural Similarity Index (SSIM) 
to measure fidelity to the ground truth, and Learned Perceptual Image Patch Similarity (LPIPS) 
to evaluate perceptual similarity.

\paragraph{Experimental settings:}
\label{setting}
We evaluate six inverse problems: denoising, box inpainting, paintbrush inpainting,
random inpainting, Gaussian deblurring, and super-resolution. 
Gaussian noise with $\sigma=0.3$ is used for denoising, $\sigma=0.05$ for box inpainting,
paintbrush inpainting, super-resolution, and deblurring, and $\sigma=0.01$ for
random inpainting. 
Box inpainting uses a centered square mask of size $40\times40$ for CelebA and
$80\times80$ for AFHQ-Cat, while random inpainting masks $70\%$ of pixels uniformly
at random and paintbrush inpainting uses irregular stochastic masks. 
Super-resolution uses downsampling factors $\times 2$ for CelebA and $\times 4$ for AFHQ-Cat. 
Gaussian deblurring uses a $61\times61$ kernel with $\sigma_b=1.0$ for CelebA and
$\sigma_b=3.0$ for AFHQ-Cat.
Subsequent work \citep{kumarpnpextended} extends the evaluation of our approach to
more challenging linear as well as nonlinear inverse problems and explores alternative
hyperparameter choices.

\paragraph{Hyperparameters:}
\label{hyperparameters}
Our ADMM-based method involves only a few hyperparameters. The main one is the penalty 
$\rho$, which controls the consensus strength between data-consistency and prior updates. 
Across all experiments in section \ref{experiments}, we fix $\rho=500$ for AFHQ-Cat and $\rho=1000$ for CelebA. 
We analyse the effect of choosing $\rho$ in an ablation study 
discussed in appendix \ref{rho_analysis}.
The data-consistency step requires solving a linear system associated with the forward operator, for which we use conjugate gradient (CG) \cite{hestenes1952methods}.
This inner solver is specified by a maximum number of 
iterations and a relative residual tolerance. 
We fix these to standard values, $\texttt{max\_cgiter}=10$ and $\texttt{cg\_tol}=10^{-5}$ in all experiments (see appendix \ref{parameters_detail}).
For a consistent evaluation, we use a fixed set of hyperparameters per dataset across all inverse tasks, without task-specific tuning. 
Likewise, each baseline uses its default hyperparameters, kept fixed across tasks.

\paragraph{Baselines:}
\label{baseline}
We benchmark our approach against recent state-of-the-art methods for linear 
inverse problems with pretrained flow-based priors: OT-ODE \cite{pokle2024training}, 
D-Flow \cite{benhamu2024dflow}, Flow Priors \cite{zhang2025flowpriors}, 
and PnP-FBS \cite{martin2025pnpflow}.

\begin{figure}[t]
    \centering
    \includegraphics[width=\linewidth]{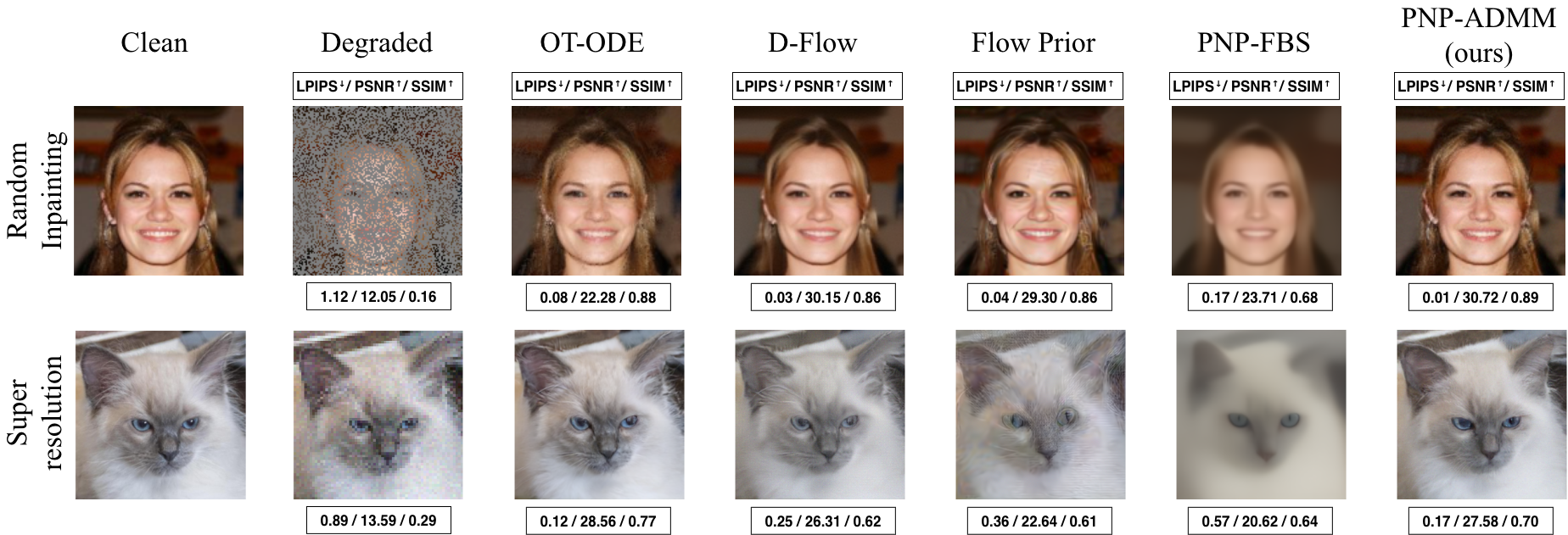}
    \caption{Comparison of image restoration methods on random inpainting and super-resolution tasks, with additional qualitative results in appendix \ref{addtional_results}. 
    Our method produces visually sharp and structurally consistent reconstructions.
    It preserves fine details while avoiding over-smoothing artifacts.}
    \label{fig:results}
\end{figure}

\begin{table*}[t]
\centering

\caption{Quantitative results on AFHQ-Cat and CelebA across various inverse tasks, 
averaged over 100 randomly selected test images per dataset. 
Best results are \textbf{bold}, second-best are shown in \second{blue}.}

\label{tab:results}

\begin{minipage}[t]{0.49\linewidth}
\centering
\setlength{\tabcolsep}{3pt}
\renewcommand{\arraystretch}{1.15}
\begin{adjustbox}{width=\linewidth}
\begin{tabular}{c l ccc ccc}
\toprule
& & \multicolumn{3}{c}{AFHQ-Cat} & \multicolumn{3}{c}{CelebA} \\
\cmidrule(lr){3-5} \cmidrule(lr){6-8}
Task & Method & SSIM$\uparrow$ & LPIPS$\downarrow$ & PSNR$\uparrow$
              & SSIM$\uparrow$ & LPIPS$\downarrow$ & PSNR$\uparrow$ \\
\midrule

\multirow{5}{*}{\rotatebox[origin=c]{90}{\makecell{Super\\Resolution}}}
& PnP-ADMM (ours)  & \second{0.64} & \second{0.16} & \second{25.19} & 0.79 & \second{0.05} & \second{28.93} \\
& OT-ODE         & \best{0.72} & \best{0.11} & \best{26.30} & \best{0.90} & \best{0.02} & \best{31.47} \\
& Flow Priors    & 0.54 & 0.32 & 21.05 & 0.80 & 0.07 & 27.17 \\
& D-Flow         & 0.58 & 0.19 & 24.33 & \second{0.82} & 0.07 & 27.36 \\
& PnP-FBS       & 0.51 & 0.55 & 18.15 & 0.70 & 0.21 & 22.47 \\
\midrule

\multirow{5}{*}{\rotatebox[origin=c]{90}{\makecell{Box\\Inpainting}}}
& PnP-ADMM (ours)  & \second{0.77} & \best{0.10} & \second{25.48} & 0.80 & \second{0.05} & \second{28.78} \\
& OT-ODE         & \best{0.87} & \best{0.10} & 23.29 & \second{0.91} & 0.06 & 27.25 \\
& Flow Priors    & \best{0.87} & \second{0.11} & \best{25.82} & \best{0.92} & \best{0.04} & 28.52 \\
& D-Flow         & 0.74 & 0.13 & 26.25 & 0.86 & 0.06 & \best{28.89} \\
& PnP-FBS       & 0.73 & 0.29 & 25.24 & 0.85 & 0.11 & 27.61 \\
\midrule

\multirow{5}{*}{\rotatebox[origin=c]{90}{\makecell{Gaussian\\Deblurring}}}
& PnP-ADMM (ours)  & 0.69 & \second{0.17} & \second{27.30} & 0.82 & 0.07 & \second{31.10} \\
& OT-ODE         & \best{0.73} & \best{0.12} & \best{27.63} & \best{0.92} & \best{0.03} & \best{32.78} \\
& Flow Priors    & \second{0.71} & 0.18 & 26.87 & \second{0.87} & \second{0.04} & 29.92 \\
& D-Flow         & 0.70 & 0.19 & 26.75 & \second{0.87} & 0.06 & 30.26 \\
& PnP-FBS       & 0.69 & 0.38 & 26.01 & 0.82 & 0.15 & 27.98 \\
\bottomrule
\end{tabular}
\end{adjustbox}
\end{minipage}%
\hfill
\begin{minipage}[t]{0.49\linewidth}
\centering
\setlength{\tabcolsep}{3pt}
\renewcommand{\arraystretch}{1.15}
\begin{adjustbox}{width=\linewidth}
\begin{tabular}{c l ccc ccc}
\toprule
& & \multicolumn{3}{c}{AFHQ-Cat} & \multicolumn{3}{c}{CelebA} \\
\cmidrule(lr){3-5} \cmidrule(lr){6-8}
Task & Method & SSIM$\uparrow$ & LPIPS$\downarrow$ & PSNR$\uparrow$
              & SSIM$\uparrow$ & LPIPS$\downarrow$ & PSNR$\uparrow$ \\
\midrule

\multirow{5}{*}{\rotatebox[origin=c]{90}{\makecell{Random\\Inpainting}}}
& PnP-ADMM (ours)  & \best{0.86} & \best{0.04} & \best{31.06} & \best{0.89} & \best{0.03} & \best{30.80} \\
& OT-ODE         & 0.81 & \second{0.09} & 29.34 & 0.84 & 0.07 & 27.78 \\
& Flow Priors    & \second{0.83} & 0.10 & \second{29.53} & 0.85 & 0.06 & 27.87 \\
& D-Flow         & 0.79 & 0.13 & 29.42 & \second{0.86} & \second{0.05} & \second{29.99} \\
& PnP-FBS       & 0.69 & 0.37 & 25.23 & 0.72 & 0.19 & 23.57 \\
\midrule

\multirow{5}{*}{\rotatebox[origin=c]{90}{Denoising}}
& PnP-ADMM (ours)  & 0.37 & 0.41 & 22.45 & 0.61 & 0.18 & 20.14 \\
& OT-ODE         & \second{0.76} & \best{0.10} & \second{28.71} & \second{0.81} & \best{0.05} & \second{28.64} \\
& Flow Priors    & 0.49 & 0.27 & 24.61 & 0.54 & 0.20 & 23.95 \\
& D-Flow         & 0.30 & 0.38 & 20.66 & 0.32 & 0.33 & 19.46 \\
& PnP-FBS       & \best{0.81} & \second{0.24} & \best{29.68} & \best{0.87} & \second{0.09} & \best{29.83} \\
\midrule

\multirow{5}{*}{\rotatebox[origin=c]{90}{\makecell{Paintbrush\\Inpainting}}}
& PnP-ADMM (ours)  & 0.77 & \second{0.10} & \second{26.74} & 0.76 & \best{0.07} & \second{25.44} \\
& OT-ODE         & \second{0.86} & 0.11 & 24.58 & \second{0.85} & 0.09 & 23.18 \\
& Flow Priors    & \best{0.89} & \best{0.09} & \best{27.67} & \best{0.86} & \second{0.08} & 24.61 \\
& D-Flow         & 0.79 & 0.12 & 26.36 & 0.82 & 0.09 & \best{26.46} \\
& PnP-FBS       & 0.79 & 0.28 & 26.57 & 0.81 & 0.14 & 25.03 \\
\bottomrule
\end{tabular}
\end{adjustbox}
\end{minipage}

\end{table*}

\paragraph{Quantitative and qualitative evaluation:}
We report reconstruction quality in table \ref{tab:results}, 
and runtime and memory in table \ref{tab:time_memory_compact}. 
Our method achieves strong performance across diverse tasks, ranking among the top methods in terms of reconstruction quality, with particularly strong results on challenging problems such as random inpainting and super-resolution. 
At the same time, it maintains low runtime and memory consumption across all settings, making it efficient and scalable in practice.
Qualitative results in fig. \ref{fig:results} show that our method yields sharp and detailed reconstructions, 
with additional results provided in appendix \ref{addtional_results}.

Our method systematically outperforms PnP-FBS across all tasks except denoising, 
and consistently ranks first or second in reconstruction quality across most inverse problems.
Denoising has an identity forward operator, making it a special case where uncertainty comes only from 
additive noise. 
In this case, closed-form methods such as Tikhonov regularization are already effective.
Nevertheless, minor adaptations of our method can further improve performance (see appendix \ref{app:denoising}).

Table \ref{tab:time_memory_compact} shows that our method is also faster and more memory-efficient than competing methods, including OT-ODE, 
while approaches such as D-Flow and Flow-Prior incur substantially higher computational costs, 
making them impractical in real-world settings.

\subsection{Analysis of Proposed Method}
\label{analysisadmm}
\begin{figure}[ht]
    \centering
    \includegraphics[width=\linewidth]{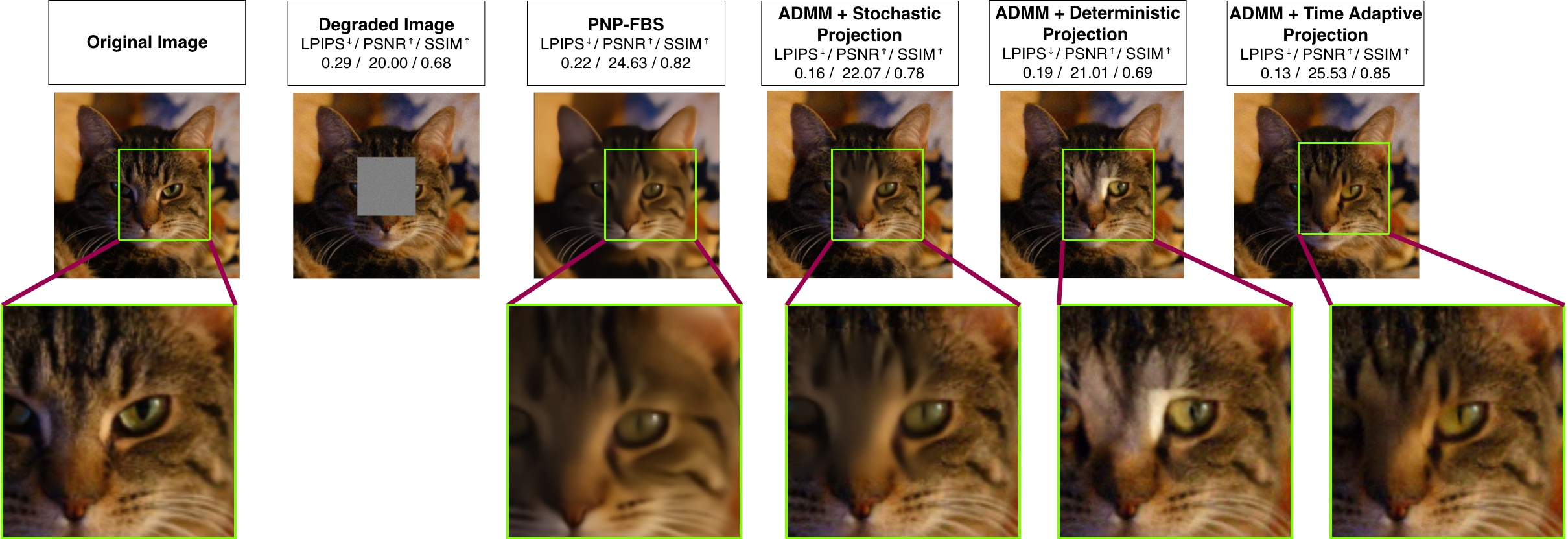}
    \caption{Comparison of update strategies for box inpainting; zoomed regions 
    highlight the inpainted area and its neighboring pixels.}
    \label{fig:ablation_cat}
\end{figure}

Fig. \ref{fig:ablation_cat} provides a qualitative comparison of different update strategies for the box inpainting, with a more detailed analysis provided in appendix~\ref{ablation}. 
For reference, we also include PnP-FBS, which produces overly smooth 
reconstructions. This behavior stems from the repeated injection of random noise throughout the 
iterations, which leads to loss of high-frequency details.

\textbf{ADMM with stochastic projections:}
Replacing FBS with ADMM significantly improves reconstruction quality by 
better preserving fine structures. 
However, this approach still yields blurry results in the inpainting region due to projections
using purely random component $\rvepsilon_R$.
While stochasticity enables exploration, it lacks guidance from the current iterate, resulting in 
suboptimal refinement of missing regions.

\textbf{ADMM with deterministic projections:}
Replacing $\rvepsilon_R$ with the deterministic component $\rvepsilon_P$ improves sharpness, as updates 
remain aligned with the current trajectory. 
However, relying solely on this strategy introduces a critical limitation: the method 
becomes highly dependent on initialization and early iterates. 
Consequently, it restricts exploration 
and may converge to suboptimal solutions. 
This reflects the inherent trade-off where trajectory-consistent 
updates bias the algorithm toward a single solution path, limiting the ability to discover alternatives 
that better satisfy the measurement constraints.

\textbf{ADMM with time adaptive projections:}
Finally, our proposed method effectively addresses this limitation through the 
time adaptive projection, in which $\rvepsilon^k$ is estimated from equation \eqref{eq:adaptive_projection}. 
By promoting exploration in the early iterations and gradually transitioning toward trajectory-consistent updates, 
the method achieves a principled balance between exploration and exploitation. 
This results in reconstructions that are both sharp and semantically consistent, 
particularly in challenging regions such as the inpainting mask.

\begin{table*}[t]

\caption{
Per-batch runtime (in seconds) and peak GPU memory (in GB) on AFHQ-Cat and CelebA 
(batch size of 4 images, averaged over 25 batches). 
Entries with the best time–memory trade-offs based on joint 
consideration of runtime and peak memory are highlighted.   
}

\label{tab:time_memory_compact}
\centering
\scriptsize

\makebox[\textwidth][c]{%
\begin{minipage}[t]{0.48\textwidth}
\centering
\setlength{\tabcolsep}{4pt}
\renewcommand{\arraystretch}{0.9}
\begin{adjustbox}{width=\linewidth}
\begin{tabular}{c l cc cc}
\toprule
& & \multicolumn{2}{c}{AFHQ-Cat} & \multicolumn{2}{c}{CelebA} \\
\cmidrule(lr){3-4} \cmidrule(lr){5-6}
Task & Method & Time$\downarrow$ & Mem.$\downarrow$
              & Time$\downarrow$ & Mem.$\downarrow$ \\
\midrule

\multirow{5}{*}{\rotatebox[origin=c]{90}{\makecell{Super\\Resolution}}}
& PnP-ADMM (ours) & \pairbest{12.81} & \pairbest{1.79} & \pairbest{7.46} & \pairbest{0.55} \\
& OT-ODE        & 27.63 & 9.52 & 16.57 & 2.63 \\
& Flow Priors   & 75.59 & 34.21 & 35.13 & 9.04 \\
& D-Flow        & 85.80 & 51.37 & 66.12 & 13.59 \\
& PnP-FBS       & 19.68 & 1.77 & 15.61 & 0.55 \\

\midrule

\multirow{5}{*}{\rotatebox[origin=c]{90}{\makecell{Box\\Inpainting}}}
& PnP-ADMM (ours) & \pairbest{10.32} & \pairbest{0.79} & \pairbest{6.59} & \pairbest{0.30} \\
& OT-ODE        & 16.34 & 8.47 & 15.03 & 2.32 \\
& Flow Priors   & 75.46 & 33.21 & 35.22 & 8.79 \\
& D-Flow        & 85.90 & 50.38 & 67.99 & 13.34 \\
& PnP-FBS       & 19.51 & 0.77 & 15.57 & 0.30 \\

\midrule

\multirow{5}{*}{\rotatebox[origin=c]{90}{\makecell{Gaussian\\Deblurring}}}
& PnP-ADMM (ours) & \pairbest{10.80} & \pairbest{0.79} & \pairbest{6.89} & \pairbest{0.30} \\
& OT-ODE        & 16.42 & 8.49 & 14.12 & 2.33 \\
& Flow Priors   & 75.55 & 33.22 & 35.86 & 8.80 \\
& D-Flow        & 85.91 & 50.38 & 68.28 & 13.34 \\
& PnP-FBS       & 19.54 & 0.78 & 15.91 & 0.30 \\

\bottomrule
\end{tabular}
\end{adjustbox}
\end{minipage}\hspace{0.02\textwidth}%
\begin{minipage}[t]{0.48\textwidth}
\centering
\setlength{\tabcolsep}{4pt}
\renewcommand{\arraystretch}{0.9}
\begin{adjustbox}{width=\linewidth}
\begin{tabular}{c l cc cc}
\toprule
& & \multicolumn{2}{c}{AFHQ-Cat} & \multicolumn{2}{c}{CelebA} \\
\cmidrule(lr){3-4} \cmidrule(lr){5-6}
Task & Method & Time$\downarrow$ & Mem.$\downarrow$
              & Time$\downarrow$ & Mem.$\downarrow$ \\
\midrule

\multirow{5}{*}{\rotatebox[origin=c]{90}{\makecell{Random\\Inpainting}}}
& PnP-ADMM (ours) & 39.58 & 0.79 & \pairbest{12.52} & \pairbest{0.30} \\
& OT-ODE        & 18.71 & 8.47 & 14.72 & 2.32 \\
& Flow Priors   & 76.14 & 33.22 & 36.18 & 8.79 \\
& D-Flow        & 86.02 & 50.37 & 66.70 & 13.34 \\
& PnP-FBS       & \pairbest{20.44} & \pairbest{0.77} & 15.63 & 0.30 \\

\midrule

\multirow{5}{*}{\rotatebox[origin=c]{90}{Denoising}}
& PnP-ADMM (ours) & \pairbest{9.95} & \pairbest{0.79} & \pairbest{6.46} & \pairbest{0.30} \\
& OT-ODE        & 16.41 & 8.47 & 14.37 & 2.32 \\
& Flow Priors   & 75.51 & 33.21 & 35.60 & 8.79 \\
& D-Flow        & 85.96 & 50.38 & 66.73 & 13.34 \\
& PnP-FBS       & 19.50 & 0.77 & 15.45 & 0.30 \\

\midrule

\multirow{5}{*}{\rotatebox[origin=c]{90}{\makecell{Paintbrush\\Inpainting}}}
& PnP-ADMM (ours) & \pairbest{12.52} & \pairbest{0.79} & \pairbest{7.90} & \pairbest{0.30} \\
& OT-ODE        & 16.99 & 8.47 & 14.42 & 2.32 \\
& Flow Priors   & 76.15 & 33.21 & 36.76 & 8.79 \\
& D-Flow        & 85.97 & 50.38 & 65.90 & 13.34 \\
& PnP-FBS       & 19.67 & 0.77 & 15.67 & 0.30 \\

\bottomrule
\end{tabular}
\end{adjustbox}
\end{minipage}%
}
\end{table*}

\subsection{Computational Complexity and Efficiency}
\label{computation}
In this section, we analyze the computational efficiency and memory footprint of all methods. 
Experiments are conducted on a single NVIDIA A100-SXM4-80GB GPU (CUDA 13.0), using a batch size 
of 4 images and averaged over 25 batches. 
Table \ref{tab:time_memory_compact} reports per-batch runtime and peak GPU memory usage.

D-Flow and Flow Priors incur significant computational and memory 
overhead, limiting practical efficiency.
D-Flow is expensive because it repeatedly solves the full ODE and backpropagates through 
the entire trajectory to optimize the initial noise. This leads to high memory usage 
from storing intermediate activations and long runtime due to repeated 
full-trajectory differentiation.
Flow Priors is computationally expensive, as each ODE step involves multiple gradient updates, requiring repeated forward and backward passes through the model along with costly Jacobian trace evaluations. This nested optimization leads to high overall runtime.
OT-ODE is more efficient as it avoids full trajectory backpropagation, 
instead applying a data-consistency correction at each ODE step that 
requires a single vector-Jacobian product (VJP) through the denoiser. 
However, computing this VJP still incurs notable memory overhead, as 
intermediate activations must be stored within each step.

In contrast, PnP-Flow methods are significantly more memory efficient. 
However, PnP-FBS in~\cite{martin2025pnpflow} uses Monte Carlo averaging in the 
denoising step to improve reconstruction quality, which increases computational 
cost. 
Specifically, PnP-FBS requires $M$ denoiser forward passes per iteration, 
leading to a complexity of $\mathcal{O}(T M C_{\mathrm{model}})$, where $T$ 
denotes the number of iterations and $C_{\mathrm{model}}$ the cost of a single 
denoiser forward pass. In contrast, our ADMM-based formulation uses a single 
denoiser evaluation per iteration, yielding $\mathcal{O}(T C_{\mathrm{model}})$, 
together with an additional CG solve for data consistency. 
In practice, the CG step is cheaper than repeated network evaluations, so runtime 
is dominated by denoiser calls, resulting in an approximate $\mathcal{O}(M)$ 
speedup over PnP-FBS. 
Both methods exhibit similarly low memory consumption, while PnP-ADMM achieves 
faster runtimes across all tasks except random inpainting in AFHQ-Cat dataset.

\section{Conclusion and Limitations}
\label{conclusion}

We presented an ADMM-based PnP framework for inverse problems with flow matching priors, 
addressing the mismatch between data-consistency updates and flow trajectories. By introducing a 
time-adaptive trajectory correction, our method effectively balances exploration and trajectory alignment, 
leading to sharper reconstructions across a range of inverse tasks.
Our experiments show that the proposed approach outperforms PnP-FBS in most settings and is competitive 
with state-of-the-art methods, while achieving significantly lower computational cost, runtime, and memory 
usage. 
These results highlight the practical advantages of PnP-ADMM with flow priors for scalable and 
efficient inverse problem solving.

A limitation of our work is the lack of a rigorous theoretical convergence analysis for PnP-ADMM with 
flow-based priors.
While we observe stable convergence behavior empirically, establishing formal guarantees 
remains challenging due to the implicit, nonlinear, and nonconvex nature of learned denoisers 
(see appendix~\ref{theory_limits}). Addressing these theoretical questions is an important direction for 
future work.

\begin{ack}
This research is supported by the Center for Artificial Intelligence and Robotics
(CAIRO) at the Technical University of Applied Sciences W\"urzburg-Schweinfurt (THWS),
W\"urzburg, Germany.

The authors have no competing interests to declare that are relevant to the content
of this article.
\end{ack}

\bibliographystyle{plainnat}
\bibliography{references}

@inproceedings{venkatakrishnan2013plug,
  title={Plug-and-play priors for model based reconstruction},
  author={Venkatakrishnan, Singanallur V and Bouman, Charles A and Wohlberg, Brendt},
  booktitle={2013 IEEE Global Conference on Signal and Information Processing},
  year={2013},
  organization={IEEE}
}

@inproceedings{ryu2019plugandplay,
      title={Plug-and-Play Methods Provably Converge with Properly Trained Denoisers}, 
      author={Ernest K. Ryu and Jialin Liu and Sicheng Wang and Xiaohan Chen and Zhangyang Wang and Wotao Yin},
      booktitle={International Conference on Machine Learning},
      eprint={arXiv:1905.05406},
      year={2019},
}

@article{Zhang_2017,
   title={Beyond a Gaussian Denoiser: Residual Learning of Deep CNN for Image Denoising},
   journal={IEEE Transactions on Image Processing},
   publisher={Institute of Electrical and Electronics Engineers},
   author={Zhang, Kai and Zuo, Wangmeng and Chen, Yunjin and Meng, Deyu and Zhang, Lei},
   year={2017},
}

@inproceedings{bora2017compressed,
      title={Compressed Sensing using Generative Models}, 
      author={Ashish Bora and Ajil Jalal and Eric Price and Alexandros G. Dimakis},
      booktitle={International Conference on Machine Learning},
      year={2017},
      eprint={arXiv:1703.03208}
}

@inproceedings{zhu2023denoising,
      title={Denoising Diffusion Models for Plug-and-Play Image Restoration}, 
      author={Yuanzhi Zhu and Kai Zhang and Jingyun Liang and Jiezhang Cao and Bihan Wen and Radu Timofte and Luc Van Gool},
      booktitle={IEEE/CVF Conference on Computer Vision and Pattern Recognition},
      year={2023},
      eprint={arXiv:2305.08995}
}

@inproceedings{mardani2023variational,
      title={A Variational Perspective on Solving Inverse Problems with Diffusion Models}, 
      author={Morteza Mardani and Jiaming Song and Jan Kautz and Arash Vahdat},
      booktitle={International Conference on Learning Representations},
      year={2024},
      eprint={arXiv:2305.04391}
}

@article{li2025decoupled,
      title={Decoupled Data Consistency with Diffusion Purification for Image Restoration}, 
      author={Xiang Li and Soo Min Kwon and Shijun Liang and Ismail R. Alkhouri and Saiprasad Ravishankar and Qing Qu},
      journal={arXiv:2403.06054},
      year={2025},
}

@inproceedings{martin2025pnpflow,
      title={PnP-Flow: Plug-and-Play Image Restoration with Flow Matching}, 
      author={Ségolène Martin and Anne Gagneux and Paul Hagemann and Gabriele Steidl},
      booktitle={International Conference on Learning Representations},
      year={2025},
      eprint={arXiv:2410.02423}
}

@inproceedings{pourya2026flower,
      title={{FLOWER}: A Flow-Matching Solver for Inverse Problems},
      author={Mehrsa Pourya and Bassam El Rawas and Michael Unser},
      booktitle={International Conference on Learning Representations},
      year={2026}
}

@inproceedings{chung2024diffusion,
      title={Diffusion Posterior Sampling for General Noisy Inverse Problems}, 
      author={Hyungjin Chung and Jeongsol Kim and Michael T. Mccann and Marc L. Klasky and Jong Chul Ye},
      year={2023},
      booktitle={International Conference on Learning Representations},
      eprint={arXiv:2209.14687}
}

@inproceedings{song2022solvinginverse,
      title={Solving Inverse Problems in Medical Imaging with Score-Based Generative Models}, 
      author={Yang Song and Liyue Shen and Lei Xing and Stefano Ermon},
      booktitle={International Conference on Learning Representations},
      year={2022},
      eprint={arXiv:2111.08005}
}

@inproceedings{li2025imagerestorationprimaldual,
      title={Image Restoration via Primal Dual Hybrid Gradient and Flow Generative Model}, 
      author={Ji Li and Chao Wang},
      booktitle={Proceedings of the AAAI Conference on Artificial Intelligence},
      year={2026},
      eprint={arXiv:2511.06748} 
}

@article{wickremasinghe2025flow,
      title={FlowSteer: Conditioning Flow Field for Consistent Image Restoration}, 
      author={Tharindu Wickremasinghe and Chenyang Qi and Harshana Weligampola and Zhengzhong Tu and Stanley H. Chan},
      year={2025},
      journal={arXiv:2512.08125}
}

@inproceedings{lipman2023flowmatching,
      title={Flow Matching for Generative Modeling}, 
      author={Yaron Lipman and Ricky T. Q. Chen and Heli Ben-Hamu and Maximilian Nickel and Matt Le},
      booktitle={International Conference on Learning Representations},
      year={2023},
      eprint={arXiv:2210.02747}
}

@inproceedings{zhang2017learning,
      title={Learning Deep CNN Denoiser Prior for Image Restoration}, 
      author={Kai Zhang and Wangmeng Zuo and Shuhang Gu and Lei Zhang},
      booktitle={IEEE Conference on Computer Vision and Pattern Recognition},
      year={2017},
      eprint={arXiv:1704.03264}
}

@article{romano2017little,
  title={The Little Engine that Could: Regularization by Denoising (RED)}, 
  author={Yaniv Romano and Michael Elad and Peyman Milanfar},
  journal={SIAM Journal on Imaging Sciences},
  year={2017}
}

@article{boyd_distributed_2010,
	title = {Distributed {Optimization} and {Statistical} {Learning} via the {Alternating} {Direction} {Method} of {Multipliers}},
	journal = {Foundations and Trends in Machine Learning},
	author = {Boyd, Stephen and Parikh, Neal and Chu, Eric and Peleato, Borja and Eckstein, Jonathan},
	year = {2010}
}

@inproceedings{yang2015facialparts,
      title={From Facial Parts Responses to Face Detection: A Deep Learning Approach}, 
      author={Shuo Yang and Ping Luo and Chen Change Loy and Xiaoou Tang},
      booktitle={IEEE International Conference on Computer Vision},
      year={2015},
      eprint={arXiv:1509.06451}
}

@inproceedings{choi2020starganv,
      title={StarGAN v2: Diverse Image Synthesis for Multiple Domains}, 
      author={Yunjey Choi and Youngjung Uh and Jaejun Yoo and Jung-Woo Ha},
      year={2020},
      booktitle={IEEE/CVF Conference on Computer Vision and Pattern Recognition},
      eprint={arXiv:1912.01865}
}

@article{pokle2024training,
      title={Training-free Linear Image Inverses via Flows},
      author={Ashwini Pokle and Matthew J. Muckley and Ricky T. Q. Chen and Brian Karrer},
      journal={Transactions on Machine Learning Research},
      year={2024},
      eprint={2310.04432},
      archivePrefix={arXiv}
}

@inproceedings{benhamu2024dflow,
      title={D-Flow: Differentiating through Flows for Controlled Generation}, 
      author={Heli Ben-Hamu and Omri Puny and Itai Gat and Brian Karrer and Uriel Singer and Yaron Lipman},
      booktitle={International Conference on Machine Learning},
      year={2024},
      eprint={arXiv:2402.14017}
}

@inproceedings{zhang2025flowpriors,
      title={Flow Priors for Linear Inverse Problems via Iterative Corrupted Trajectory Matching}, 
      author={Yasi Zhang and Peiyu Yu and Yaxuan Zhu and Yingshan Chang and Feng Gao and Ying Nian Wu and Oscar Leong},
      booktitle={Advances in Neural Information Processing Systems},
      year={2024},
      eprint={arXiv:2405.18816}
}

@inproceedings{chung2024improving,
      title={Improving Diffusion Models for Inverse Problems using Manifold Constraints}, 
      author={Hyungjin Chung and Byeongsu Sim and Dohoon Ryu and Jong Chul Ye},
      booktitle={Advances in Neural Information Processing Systems},
      year={2022},
      eprint={arXiv:2206.00941},
}

@inproceedings{song2023pseudo,
title={Pseudoinverse-Guided Diffusion Models for Inverse Problems},
author={Jiaming Song and Arash Vahdat and Morteza Mardani and Jan Kautz},
booktitle={International Conference on Learning Representations},
year={2023}
}

@article{hestenes1952methods,
  author = {Hestenes, Magnus R and Stiefel, Eduard},
  journal = {Journal of research of the National Bureau of Standards},
  title = {Methods of conjugate gradients for solving linear systems},
  year = 1952
}

@article{chan2016plug,
      title={Plug-and-Play ADMM for Image Restoration: Fixed Point Convergence and Applications}, 
      author={Stanley H. Chan and Xiran Wang and Omar A. Elgendy},
      journal={IEEE Transactions on Computational Imaging},
      eprint={arXiv:1605.01710},
      year={2016},
}

@inproceedings{shrestha2026,
      title={Taming Score-Based Denoisers in ADMM: A Convergent Plug-and-Play Framework}, 
      author={Rajesh Shrestha and Xiao Fu},
      booktitle={International Conference on Learning Representations},
      eprint={arXiv:2603.10281},
      year={2026},
}

@inproceedings{hurault2022gradient,
  author    = {Hurault, Samuel and Leclaire, Arthur and Papadakis, Nicolas},
  title     = {Gradient Step Denoiser for Convergent Plug-and-Play},
  booktitle={International Conference on Learning Representations},
  year = {2022}
}

@inproceedings{Blau_2018,
   title={The Perception-Distortion Tradeoff},
   booktitle={2018 IEEE/CVF Conference on Computer Vision and Pattern Recognition},
   publisher={IEEE},
   author={Blau, Yochai and Michaeli, Tomer},
   year={2018}
}

@inproceedings{liu2022flowstraight,
      title={Flow Straight and Fast: Learning to Generate and Transfer Data with Rectified Flow}, 
      author={Xingchao Liu and Chengyue Gong and Qiang Liu},
      booktitle={International Conference on Learning Representations},
      eprint={arXiv:2209.03003},
      year={2023},
}

@book{Bertero1998Introduction,
  title={Introduction to Inverse Problems in Imaging},
  author={Mario Bertero and Patrizia Boccacci},
  publisher={Institute of Physics Publishing},
  year={1998},
}

@book{Tikhonov1977Solutions,
  title={Solutions of ill-posed problems},
  author={A. N. Tikhonov and Vasiliy Yakovlevich Arsenin},
  publisher={Winston \& Sons},
  year={1977},
}

@book{vogel2002computational,
author = {Vogel, Curtis R.},
title = {Computational Methods for Inverse Problems},
publisher = {Society for Industrial and Applied Mathematics},
year = {2002},
}

@article{rudin1992nonlinear,
  author  = {Rudin, Leonid I. and Osher, Stanley and Fatemi, Emad},
  title   = {Nonlinear Total Variation based Noise Removal Algorithms},
  journal = {Physica D: Nonlinear Phenomena},
  year    = {1992}
}

@book{boyd2004convex,
  title={Convex Optimization},
  author={Boyd, Stephen and Vandenberghe, Lieven},
  publisher={Cambridge University Press},
  year={2004},
}

@article{kumarpnpextended,
      title={PnP-CF: A Consensus Flow-Matching Solver for General Inverse Problems}, 
      author={Dibyanshu Kumar and Magda Gregorova},
      year={2026},
      eprint={2610.26000},
      archivePrefix={arXiv},
      primaryClass={cs.LG},
      url={https://arxiv.org/abs/2610.26000}, 
}

\clearpage
\appendix

\section{Convergence Discussion and Empirical Evidence}
\label{app:conv}

\subsection{Theoretical Background and Limitations}
\label{theory_limits}

For classical ADMM, convergence follows from convexity and proximal structure of the
subproblems \cite{boyd_distributed_2010}. 
In PnP-ADMM, however, the prior step is
replaced by a learned denoiser that does not in general correspond to the proximal 
operator of any explicit regularizer, making the regularizer implicit and rendering 
the standard convex ADMM theory inapplicable. 
As a result, existing PnP analyses typically
establish \emph{fixed-point convergence} rather than convergence to a minimizer of an
explicit objective \cite{chan2016plug,ryu2019plugandplay}.

To obtain a rigorous fixed-point convergence result for the present ADMM-flow scheme,
one would need to establish suitable operator-theoretic properties of the denoiser
$\mathcal{D}_t$. 
In the PnP-ADMM literature, such guarantees are commonly derived under
assumptions such as nonexpansiveness or averagedness of the
denoiser (or of a related residual operator), which in turn imply that the reduced
Douglas-Rachford / ADMM operator is averaged and therefore convergent to a fixed
point \cite{chan2016plug,ryu2019plugandplay}. 
More recent work shows that stronger
results, such as stationary-point convergence, can be obtained for more structured
denoisers, for example gradient-step denoisers or specially trained conservative
operators, but such guarantees do not directly extend to general generative priors
or flow-matching denoisers \cite{hurault2022gradient}. 
The recent score-based ADMM
framework of \cite{shrestha2026} follows the same philosophy: it derives convergence by
proving weak nonexpansiveness or boundedness properties of the denoiser under additional
structural assumptions.

Our denoiser has the form
\[
\mathcal{D}_t(\rvx)=\rvx+(1-t)v_t(\rvx),
\]
where $v_t$ is the learned flow-matching velocity field. To analyze its stability,
consider two inputs $\rvx_1,\rvx_2 \in \mathbb{R}^m$. Then
\begin{align}
\mathcal{D}_t(\rvx_1) - \mathcal{D}_t(\rvx_2)
&= \rvx_1 - \rvx_2 + (1-t)\bigl(v_t(\rvx_1) - v_t(\rvx_2)\bigr).
\end{align}
Taking the $\ell_2$ norm and applying the triangle inequality yields
\begin{align}
\|\mathcal{D}_t(\rvx_1) - \mathcal{D}_t(\rvx_2)\|_2
&\le \|\rvx_1 - \rvx_2\|_2 
+ (1-t)\|v_t(\rvx_1) - v_t(\rvx_2)\|_2.
\end{align}
Assuming that $v_t$ is $L_t$-Lipschitz, i.e.,
\[
\|v_t(\rvx_1) - v_t(\rvx_2)\|_2 \le L_t \|\rvx_1 - \rvx_2\|_2,
\]
we obtain
\begin{align}
\|\mathcal{D}_t(\rvx_1) - \mathcal{D}_t(\rvx_2)\|_2
&\le \|\rvx_1 - \rvx_2\|_2 
+ (1-t)L_t \|\rvx_1 - \rvx_2\|_2 \\
&= \bigl(1 + (1-t)L_t\bigr)\|\rvx_1 - \rvx_2\|_2.
\end{align}
For $\mathcal{D}_t$ to be nonexpansive, one would require
\[
\|\mathcal{D}_t(\rvx_1) - \mathcal{D}_t(\rvx_2)\|_2
\le \|\rvx_1 - \rvx_2\|_2,
\]
which implies $1 + (1-t)L_t \le 1$. However, for any $L_t > 0$ and $t < 1$,
\[
1 + (1-t)L_t > 1,
\]
so $\mathcal{D}_t$ is generally \emph{expansive}. Therefore, plain Lipschitz continuity
of $v_t$ is not sufficient to guarantee nonexpansiveness (or firm nonexpansiveness)
of $\mathcal{D}_t$.

Thus, a full ADMM convergence proof would require additional structure on $v_t$,
such as monotonicity, cocoercivity, or architectural constraints ensuring that
$\mathcal{D}_t$ is an averaged operator. Establishing such properties for general
flow-matching networks is beyond the scope of the present work and remains an
interesting direction for future research.

For this reason, we do not claim a full theoretical convergence proof for the complete
time-varying ADMM-flow algorithm. 
Instead, we provide empirical convergence diagnostics
that assess whether the iterates stabilize in practice and whether the ADMM consensus
constraint is progressively enforced.

\subsection{Empirical convergence diagnostics}
\label{empirical_convergence}

\begin{figure}[ht]
    \centering
    \includegraphics[width=\linewidth]{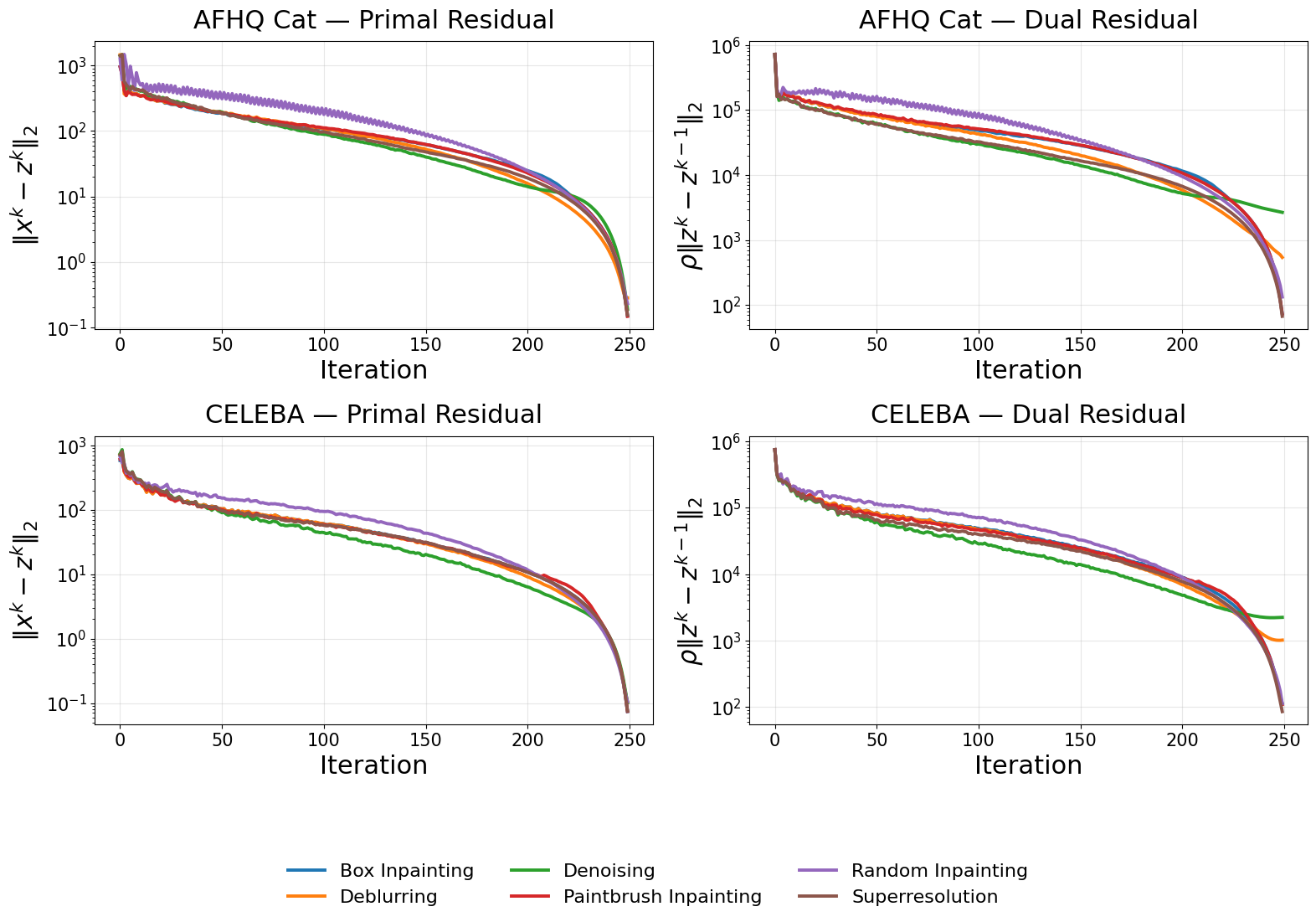}
    \caption{Empirical convergence diagnostics of the proposed PnP-Flow-ADMM method across
multiple inverse problems on the AFHQ-CAT (top) and CelebA (bottom)
datasets. 
We plot the primal residual $\|\rvx^k-\rvz^k\|_2$ (left) and the dual
residual $\rho\|\rvz^k-\rvz^{k-1}\|_2$ (right) as a function of iterations.
Across all tasks, the residuals decrease steadily, indicating progressive enforcement
of the ADMM consensus constraint and stabilization of the iterates.}
\label{fig:convergence_analysis}
\end{figure}

We evaluate convergence using standard ADMM diagnostics. 

The \emph{primal residual}
\[
r_{\mathrm{pri}}^k := \|\rvx^k - \rvz^k\|_2
\]
measures violation of the consensus constraint $\rvx=\rvz$. Its decay indicates that
the data-consistency variable and prior variable agree.

The \emph{dual residual}
\[
r_{\mathrm{dual}}^k := \rho \|\rvz^k - \rvz^{k-1}\|_2
\]
measures the change in the auxiliary variable across iterations and reflects
stabilization of the dual updates. 
In standard ADMM, convergence is characterized by
the joint decay of both primal and dual residuals \cite{boyd_distributed_2010}.

We evaluate these diagnostics across multiple inverse problems on the AFHQ and 
CelebA datasets. 
The same network,
degradation models and penalty parameter $\rho$ as in the
main experiments are used. 
At each iteration, we record the residuals and plot them
against the iteration index $k$.

\subsection{Interpretation of the plots}
\label{interpret}

We show the evolution of the primal and
dual residuals across different tasks and datasets in figure ~\ref{fig:convergence_analysis}.
The primal residual $\|\rvx^k-\rvz^k\|_2$ decreases steadily for all the tasks.
This indicates that the ADMM consensus constraint is progressively enforced. 
The dual residual $\rho\|\rvz^k-\rvz^{k-1}\|_2$ also decreases, 
showing that updates to the auxiliary
variable diminish and the iterates approach a stable regime.

We observe that \texttt{denoising} and, to a lesser extent, \texttt{deblurring} exhibit 
strong primal convergence, yet their dual residuals show slower decay toward the end 
of the iterations compared to other tasks, which undergo a sharper final drop. 
This suggests 
that although the consensus constraint is well enforced, the auxiliary variable 
updates do not fully stabilize. 
While this behavior has limited impact on the 
reconstruction quality for deblurring, it affects denoising more noticeably; 
in section \ref{admm_adaptation}, we introduce minor adaptations to mitigate this effect.

Overall, the consistent decay of both primal and dual residuals across datasets and
tasks provides strong empirical evidence that the proposed ADMM-flow iteration
stabilizes and operates in a fixed-point regime in practice.

\section{Data Consistency via Conjugate Gradient}
\label{appendix:cgd}

\subsection{Data-Consistency Subproblem}

For Gaussian noise with variance $\sigma^2$, the data-fidelity term is
\begin{equation}
    l(\rvy;\rvx) = \frac{1}{2\sigma^2}\|\mathcal{A}\rvx - \rvy\|_2^2.
\end{equation}
Substituting into the $\rvx$-update \eqref{eq:x_update} gives the quadratic subproblem
\begin{equation}
    \rvx^{k+1}
    = \arg\min_{\rvx} \;
    \frac{1}{2\sigma^2}\|\mathcal{A}\rvx - \rvy\|_2^2
    + \frac{\rho}{2}\|\rvx - (\rvz^k - \rvu^k)\|_2^2,
    \label{eq:x_subproblem_cg}
\end{equation}
where the first term enforces measurement consistency and the second anchors
$\rvx$ to the current ADMM estimate $\rvz^k - \rvu^k$.

\subsection{Reduction to a Linear System}
Since \eqref{eq:x_subproblem_cg} is a strongly convex quadratic, its unique
minimizer satisfies the first-order optimality condition
\begin{equation}
\nabla_{\rvx}\left[
\frac{1}{2\sigma^2}\|\mathcal{A}\rvx - \rvy\|_2^2
+ \frac{\rho}{2}\|\rvx - (\rvz^k - \rvu^k)\|_2^2
\right] = \mathbf{0}
\end{equation}

Setting the gradient to zero and collecting terms yields
\begin{equation}
\underbrace{\left(
    \frac{1}{\sigma^2}\mathcal{A}^\top \mathcal{A} + \rho \mI
\right)}_{\mathcal{M}}
\rvx
=
\underbrace{\frac{1}{\sigma^2}\mathcal{A}^\top \rvy
+ \rho(\rvz^k - \rvu^k)}_{\rvb},
\label{eq:linear_system_cg}
\end{equation}
i.e.\ a linear system $\mathcal{M}\rvx = \rvb$ with operator and right-hand
side as
\begin{equation}
    \mathcal{M}(\rvx)
    = \frac{1}{\sigma^2}\mathcal{A}^\top(\mathcal{A}\rvx) + \rho\,\rvx,
    \qquad
    \rvb
    = \frac{1}{\sigma^2}\mathcal{A}^\top \rvy + \rho(\rvz^k - \rvu^k).
\end{equation}
Symmetry of $\mathcal{M}$ follows from the fact that 
$\mathcal{A}^\top\mathcal{A}$ is self-adjoint and $\rho\mI$ is symmetric.
Positive definiteness follows since
\begin{equation}
    \langle \rvx,\, \mathcal{M}(\rvx) \rangle
    = \frac{1}{\sigma^2}\|\mathcal{A}\rvx\|_2^2 + \rho\|\rvx\|_2^2 > 0
    \qquad \forall\,\rvx \neq 0.
\end{equation}
guaranteeing a unique solution and making \eqref{eq:linear_system_cg} amenable
to the conjugate gradient (CG) method.

\subsection{Conjugate Gradient Solver and Practical Details}
\label{parameters_detail}

Explicitly forming or inverting the system matrix is computationally
infeasible at the image resolutions considered here.
We therefore solve \eqref{eq:linear_system_cg} iteratively with CG 
\cite{hestenes1952methods,boyd2004convex},
which requires only matrix-vector products through $\mathcal{M}$, each
reducible to one application of the forward operator $\mathcal{A}$ and one of
its adjoint $\mathcal{A}^\top$.
CG is initialized with the previous iterate $\rvx^k$ and terminated when the
relative residual falls below a prescribed tolerance,
\begin{equation}
    \frac{\|\rvr^{(j)}\|_2}{\|\rvb\|_2 + \epsilon} < \texttt{cg\_tol},
\end{equation}
where $\rvb$ denotes the right-hand side of the linear system in 
\eqref{eq:linear_system_cg}, $\rvr^{(j)} = \rvb - \mathcal{M}\rvx^{(j)}$ 
is the residual at the $j$-th CG iteration, and $\epsilon > 0$ is a small 
constant introduced for numerical stability to avoid division by zero.
The iterations are terminated either when this criterion is satisfied or after 
a maximum of $\texttt{max\_cgiter}$ iterations.
In all experiments we use $\texttt{max\_cgiter}=10$, 
$\texttt{cg\_tol}=10^{-5}$.
The ADMM penalty is set to $\rho=1000$ for CelebA and $\rho=500$ for AFHQ-Cat.

\section{Effect of the penalty parameter \texorpdfstring{$\mathbf{\rho}$}{rho}}
\label{rho_analysis}

The penalty parameter $\rho$ governs the strength of the consensus constraint 
between the data-consistency variable $\rvx$ and the prior-driven variable $\rvz$. 
In particular, larger values of $\rho$ enforce stronger agreement between the 
two variables, effectively biasing the $\rvx$-update toward the denoised estimate 
$(\rvz - \rvu)$. As a result, $\rho$ controls the trade-off between adherence to 
the measurements and reliance on the learned prior.

\begin{figure}[ht]
    \centering
    \includegraphics[width=\linewidth]{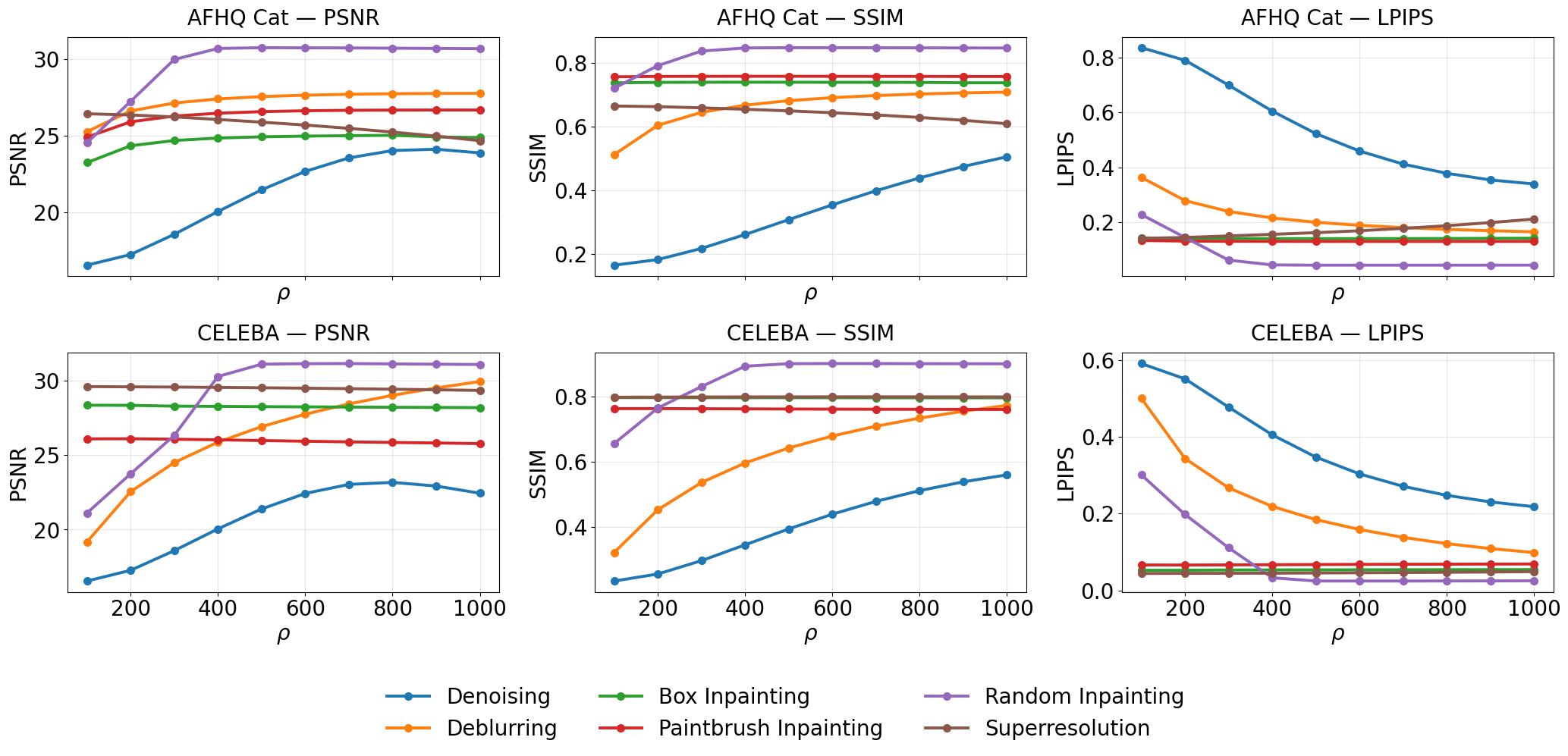}
    \caption{Effect of the penalty parameter $\rho$ on AFHQ-Cat (top) and CelebA (bottom)
    across inverse problems.
Increasing $\rho$ improves performance for denoising, deblurring, and random inpainting, 
with gains saturating at higher values, while showing marginal sensitivity for structured 
inpainting tasks and slight degradation for super-resolution.}
    \label{fig:rho_analysis}
\end{figure}

Figure~\ref{fig:rho_analysis} illustrates the impact of $\rho$ across different 
inverse problems and datasets. We observe a consistent trend for 
\texttt{denoising}, \texttt{deblurring}, and 
\texttt{random inpainting}, where increasing $\rho$ leads to substantial 
improvements in PSNR and SSIM, along with a corresponding decrease in LPIPS. 
This indicates that these tasks benefit from stronger coupling between data-consistency and the prior.
In contrast, tasks such as \texttt{box inpainting} and 
\texttt{paintbrush inpainting} exhibit only mild sensitivity to $\rho$. 
For \texttt{superresolution}, we observe a slight performance degradation as $\rho$ increases, likely because the downsampled measurements lack high-frequency information and stronger coupling, together with the perception–distortion trade-off \cite{Blau_2018}, can lead to smoother reconstructions with reduced fine detail.

Despite this, we observe that larger values of 
$\rho$ consistently yield significant improvements across the majority of tasks. 
In these settings, stronger consensus 
enforcement allows the denoiser to effectively regularize the solution and 
mitigate the ill-posedness of the inverse problem.
Based on these observations, we select $\rho$ from this regime, favoring larger 
values that provide consistent gains across tasks while maintaining stability. 
Specifically, we use $\rho = 500$ for AFHQ-Cat and $\rho = 1000$ for CelebA, 
as these values lie in the saturation region where the method achieves a 
favorable balance between data consistency and prior regularization, with only 
minor degradation in super-resolution.

\section{Ablation Studies}
\label{ablation}

\begin{figure}[ht]
\centering
\includegraphics[width=\linewidth,height=0.73\textheight,keepaspectratio]{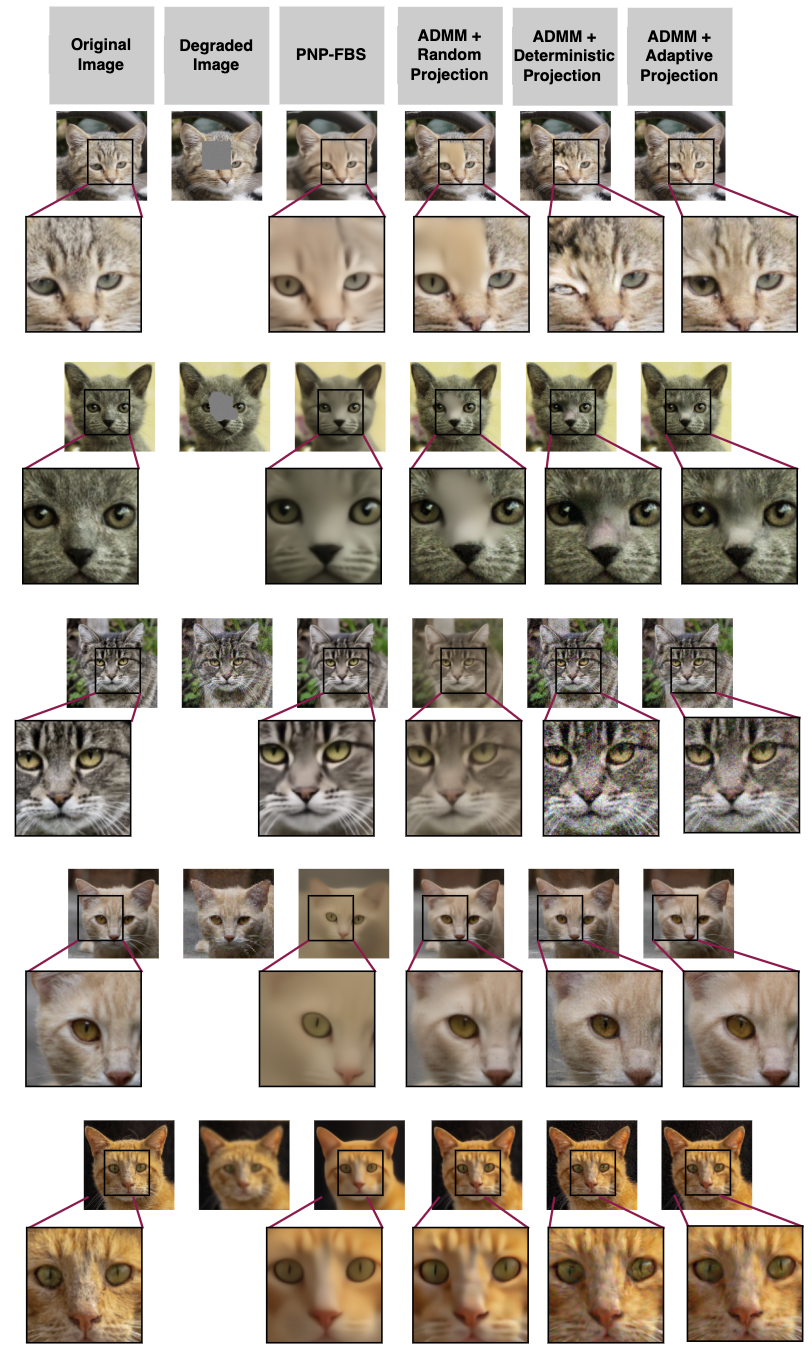}
\caption{Qualitative comparison across inverse tasks on AFHQ-Cat. Columns (left→right): original, degraded, PnP-FBS, ADMM (stochastic), ADMM (deterministic), and ADMM (adaptive). Rows (top→bottom): box inpainting, paintbrush inpainting, denoising, super-resolution, and deblurring. PnP-FBS yields overly smooth results; ADMM (stochastic) improves structure but remains blurry; ADMM (deterministic) sharpens details but introduces artifacts. The proposed adaptive projection achieves sharper, more consistent reconstructions, as highlighted in the zoomed regions.}
\label{ablation:bpsd}
\end{figure}

\begin{figure}[ht]
\centering
\includegraphics[width=\linewidth,height=0.73\textheight,keepaspectratio]{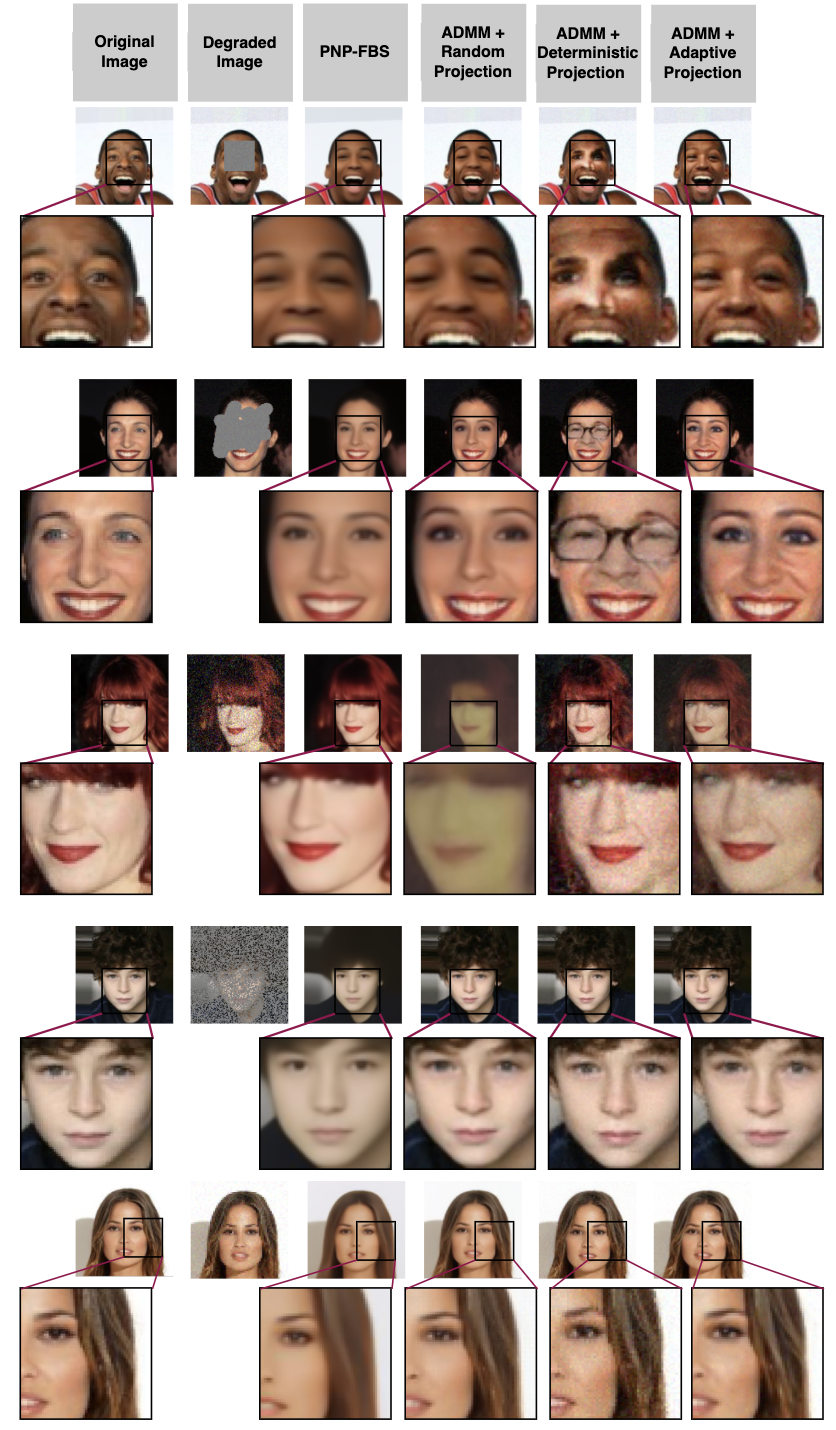}
\caption{Qualitative comparison across inverse tasks on CelebA. Columns (left→right): original, degraded, PnP-FBS, ADMM (stochastic), ADMM (deterministic), and ADMM (adaptive). Rows (top→bottom): box inpainting, paintbrush inpainting, denoising, random inpainting, and super-resolution.}
\label{ablation:celeba_bpsrd}
\end{figure}

In this section, we study the impact of different design choices in our method 
through additional ablation experiments. Specifically, we analyze how different 
projection strategies and optimization schemes affect reconstruction performance, 
providing further support for the analysis presented in 
Section~\ref{analysisadmm}. 

We consider multiple variants of our framework, in addition to the 
\textbf{PnP-FBS} baseline of \cite{martin2025pnpflow}. 
First, we evaluate \textbf{PnP-ADMM with stochastic projections}, where the projection 
step relies purely on random noise. 
Second, we study \textbf{PnP-ADMM with deterministic projections}, where only 
the deterministic component is used. 
Finally, we consider our full approach, \textbf{PnP-ADMM with time adaptive projections}, 
which combines both stochastic and deterministic components in an adaptive manner 
within the ADMM framework.

\begin{table}[ht]
\centering
\caption{Average performance of different settings across all inverse tasks (denoising,
deblurring, random inpainting, box inpainting, paintbrush inpainting and superresolution).}
\label{tab:avg_results_images}
\setlength{\tabcolsep}{4pt}
\renewcommand{\arraystretch}{1.2}
\begin{tabular}{l ccc ccc}
\toprule
& \multicolumn{3}{c}{AFHQ-CAT} & \multicolumn{3}{c}{CelebA} \\
\cmidrule(lr){2-4} \cmidrule(lr){5-7}
Method 
& SSIM$\uparrow$ & LPIPS$\downarrow$ & PSNR$\uparrow$
& SSIM$\uparrow$ & LPIPS$\downarrow$ & PSNR$\uparrow$ \\
\midrule

PNP-ADMM (adaptive)      
& \best{0.74} & \best{0.16} & \best{26.26}
& \best{0.80} & \best{0.07} & \best{27.55} \\

PNP-ADMM (random)    
& 0.71 & 0.21 & 25.81
& 0.79 & 0.09 & 26.47 \\

PNP-ADMM (deterministic) 
& 0.61 & 0.23 & 23.78
& 0.73 & 0.10 & 26.16 \\

PNP-FBS        
& 0.70 & 0.36 & 25.15
& 0.79 & 0.15 & 26.18 \\

\bottomrule
\end{tabular}
\end{table}

Table~\ref{tab:avg_results_images} reports the average performance across all inverse 
tasks.
ADMM-based variants achieve better results than the PnP-FBS baseline across most metrics,
demonstrating the benefit of the ADMM formulation over forward-backward splitting.
Among these, ADMM with time-adaptive projection achieves the best overall performance across all the metrics.

This is further corroborated by qualitative evaluation. As shown in 
fig.~\ref{ablation:bpsd} and fig~\ref{ablation:celeba_bpsrd}, PnP-FBS 
produces overly smooth reconstructions, while ADMM with random projections, 
although improved, still fails to recover fine-grained details. 
ADMM with deterministic projection yields sharper images but introduces visible artifacts. 
In contrast, the proposed adaptive projection strategy strikes a favorable balance: 
reconstructions are both perceptually sharp and free of structural artifacts.

\section{PnP-FBS Algorithm}
\label{fbs_algo}

Below we provide the PnP-FBS algorithm \cite{martin2025pnpflow} used as one of the baselines 
in our experiments. 
The method alternates between a data-consistency gradient step and a denoising step 
using a time-dependent flow-based prior, with an additional projection onto the 
intermediate flow trajectory at each iteration.

\begin{algorithm}[H]
\caption{PnP-FBS}
\label{algo:pnpflow_fbs}
\begin{algorithmic}[1]
\STATE \textbf{Input:} Denoiser $\mathcal{D}_t$, time sequence $(t_k)_k$ either 
finite with $t_k = k/N$, $N \in \mathbb{N}$, or infinite with $\lim_{k\to\infty} t_k = 1$, 
time dependent step size $\gamma(t_k)$, $p_0 = \mathcal{N}(0, I)$
\STATE \textbf{Initialize:} $\rvx^0 \in \mathbb{R}^m$
\FOR{$k = 0,1,\dots$}
    \STATE $\rvz^k = \rvx^k - \gamma(t_k) \nabla l(\rvy;\rvx^k)$
    \STATE $\tilde{\rvz}^k = (1 - t_k)\rvepsilon + t_k \rvz^k,\quad \rvepsilon \sim \mathcal{N}(0,I)$
    \STATE $\rvx^{k+1} = \mathcal{D}_{t_k}(\tilde{\rvz}^k)$
\ENDFOR
\STATE \textbf{return} $\rvx^{k+1}$
\end{algorithmic}
\end{algorithm}

\section{Denoising as a Simple Inverse Problem}
\label{app:denoising}

While our method performs strongly across most inverse problems, its performance on denoising is comparatively lower. 
In this section, we discuss what makes denoising fundamentally different, show that the problem can be effectively addressed using standard Tikhonov regularization, and demonstrate that minor adaptations of our method can further improve performance.

Denoising is a particularly simple inverse problem in which the forward 
operator \(\mathcal{A} = I\) is the identity. The observation model is
\begin{equation}
\rvy = \rvx + \rvepsilon, \quad 
\rvepsilon \sim \mathcal{N}(0, \sigma^2 I),
\end{equation}
where \(\rvx \in \mathbb{R}^n\) is the clean signal and \(\rvy\) is the 
noisy observation. Unlike ill-posed problems such as inpainting or 
super-resolution \cite{Bertero1998Introduction}, the forward operator is 
perfectly conditioned and introduces no structural loss of information. 
The remaining uncertainty arises solely from additive noise, so the 
data-fidelity term
\begin{equation}
\min_{\rvx} \frac{1}{2\sigma^2}\|\rvx - \rvy\|^2
\label{eq:data_fidelity}
\end{equation}
is already well-posed, and regularization 
serves only to suppress noise rather than to recover lost information.

\subsection{Closed-Form Solution}

A classical approach to solving the denoising problem is first-order Tikhonov 
regularization \cite{Tikhonov1977Solutions},
\begin{equation}
\min_{\rvx} \frac{1}{2\sigma^2}\|\rvx - \rvy\|^2 
+ \frac{\lambda}{2}\|\nabla \rvx\|^2,
\end{equation}
where \(\lambda > 0\) is a regularization parameter that controls the trade-off 
between data fidelity and smoothness. The optimality condition, after multiplying 
through by \(\sigma^2\), is given by
\begin{equation}
\left(I + \lambda \sigma^2 \nabla^\top \nabla \right)\rvx = \rvy.
\end{equation}
Since \(\nabla^\top \nabla\) diagonalizes under the discrete Fourier 
transform \cite{vogel2002computational}, this system admits the closed-form 
solution
\begin{equation}
\widehat{\rvx}^\star(\boldsymbol{\xi}) 
= \frac{\widehat{\rvy}(\boldsymbol{\xi})}
{1 + \lambda \sigma^2 \|\boldsymbol{\xi}\|^2},
\label{tikhonov}
\end{equation}
where \(\boldsymbol{\xi}\) denotes the spatial frequency. The solution acts 
as a low-pass filter, attenuating high-frequency noise while preserving 
low-frequency structure~\cite{rudin1992nonlinear}, and is obtained in a 
single step without iterative optimization.

\begin{table}[h]
\caption{Denoising performance using Tikhonov regularization for different values of $\lambda$ on AFHQ-Cat and CelebA datasets.
Best results are obtained around $\lambda=15$ using the closed-form solution in Eq.~\eqref{tikhonov}.}
\label{tab:tikhonov_denoising}
\centering
\small
\setlength{\tabcolsep}{5pt}
\renewcommand{\arraystretch}{1.1}

\begin{tabular}{c|ccc|ccc}
\toprule
 & \multicolumn{3}{c|}{AFHQ-Cat} & \multicolumn{3}{c}{CelebA} \\
\cmidrule(lr){2-4} \cmidrule(lr){5-7}
$\lambda$ & SSIM $\uparrow$ & LPIPS $\downarrow$ & PSNR $\uparrow$ 
          & SSIM $\uparrow$ & LPIPS $\downarrow$ & PSNR $\uparrow$ \\
\midrule
5  & 0.51 & 0.41 & 24.91 & 0.55 & 0.31 & 24.26 \\
10 & 0.56 & 0.38 & 25.73 & 0.60 & 0.29 & 24.85 \\
15 & \best{0.60} & \best{0.36} & \best{26.13} & \best{0.62} & \best{0.28} & \best{25.04} \\
20 & 0.54 & 0.39 & 25.29 & 0.58 & 0.30 & 24.45 \\
\bottomrule
\end{tabular}
\end{table}

Table~\ref{tab:tikhonov_denoising} shows that Tikhonov regularization 
achieves strong denoising performance across all metrics, with the best 
results obtained around $\lambda = 15$. This highlights that simple 
closed-form methods are already effective for denoising, providing 
high-quality reconstructions in a single step without iterative or 
sampling procedures.

\subsection{Adaptation of PnP-ADMM for Denoising}
\label{admm_adaptation}

In the convergence analysis (section \ref{interpret}), we observe that for denoising the primal residual $\|\rvx^k - \rvz^k\|_2$ decreases rapidly, indicating early consensus between the data-consistency and denoiser outputs, while the dual residual $\rho\|\rvz^{k+1} - \rvz^k\|_2$ plateaus at a non-zero value, suggesting that the denoiser continues to introduce small variations across iterations without fully stabilizing.

\begin{table}[h]
\caption{Performance of our method on denoising with different numbers of PnP steps. 
Early stopping is applied at $85\%$ of the total steps that is selected based on empirical tuning.}
\label{tab:early_stop_denoising}
\centering
\small
\setlength{\tabcolsep}{5pt}
\renewcommand{\arraystretch}{1.1}

\begin{tabular}{c c|ccc|ccc}
\toprule
\multirow{2}{*}{PnP steps} & \multirow{2}{*}{Early Stop} 
& \multicolumn{3}{c|}{AFHQ-Cat} 
& \multicolumn{3}{c}{CelebA} \\
\cmidrule(lr){3-5} \cmidrule(lr){6-8}
 & 
 & SSIM $\uparrow$ & LPIPS $\downarrow$ & PSNR $\uparrow$
 & SSIM $\uparrow$ & LPIPS $\downarrow$ & PSNR $\uparrow$ \\
\midrule
200  & 170  &0.58  &0.30  &19.05  &0.60  &0.19  &17.58  \\
300  & 255  &0.65  &0.22  &21.68  &0.67  &0.14  &19.76  \\
400  & 340  &0.68  &0.19  &23.63  &0.66  &0.12  &23.51  \\
800  & 680  &\best{0.72}  &0.13  &27.47  &\best{0.76}  &0.07  &26.01  \\
1000 & 850  &\best{0.72}  &\best{0.12}  &\best{28.05}  &\best{0.76}  &\best{0.06}  &\best{26.99}  \\
\bottomrule
\end{tabular}
\end{table}

A possible explanation is that, since the forward operator is the identity, the iterates quickly approach the data manifold, after which the denoiser operates in a regime where updates become small and less stable, leading to diminishing improvements and potential variations in perceptual quality. Motivated by this observation, we adopt a combined strategy of early stopping 
together with an increased number of Euler steps for sampling.

Table~\ref{tab:early_stop_denoising} shows that increasing the number of 
PnP steps together with early stopping consistently improves performance 
across all metrics.
This suggests that 
using more Euler steps improves the accuracy of the denoiser, while early 
stopping prevents unnecessary late-stage updates. 
Overall, this combination 
provides a simple and effective adaptation of PnP-ADMM for denoising.

\section{Additional Qualitative Results}
\label{addtional_results}
\begin{figure}[H]
    \centering
    \includegraphics[width=\textwidth,height=\textheight,keepaspectratio]{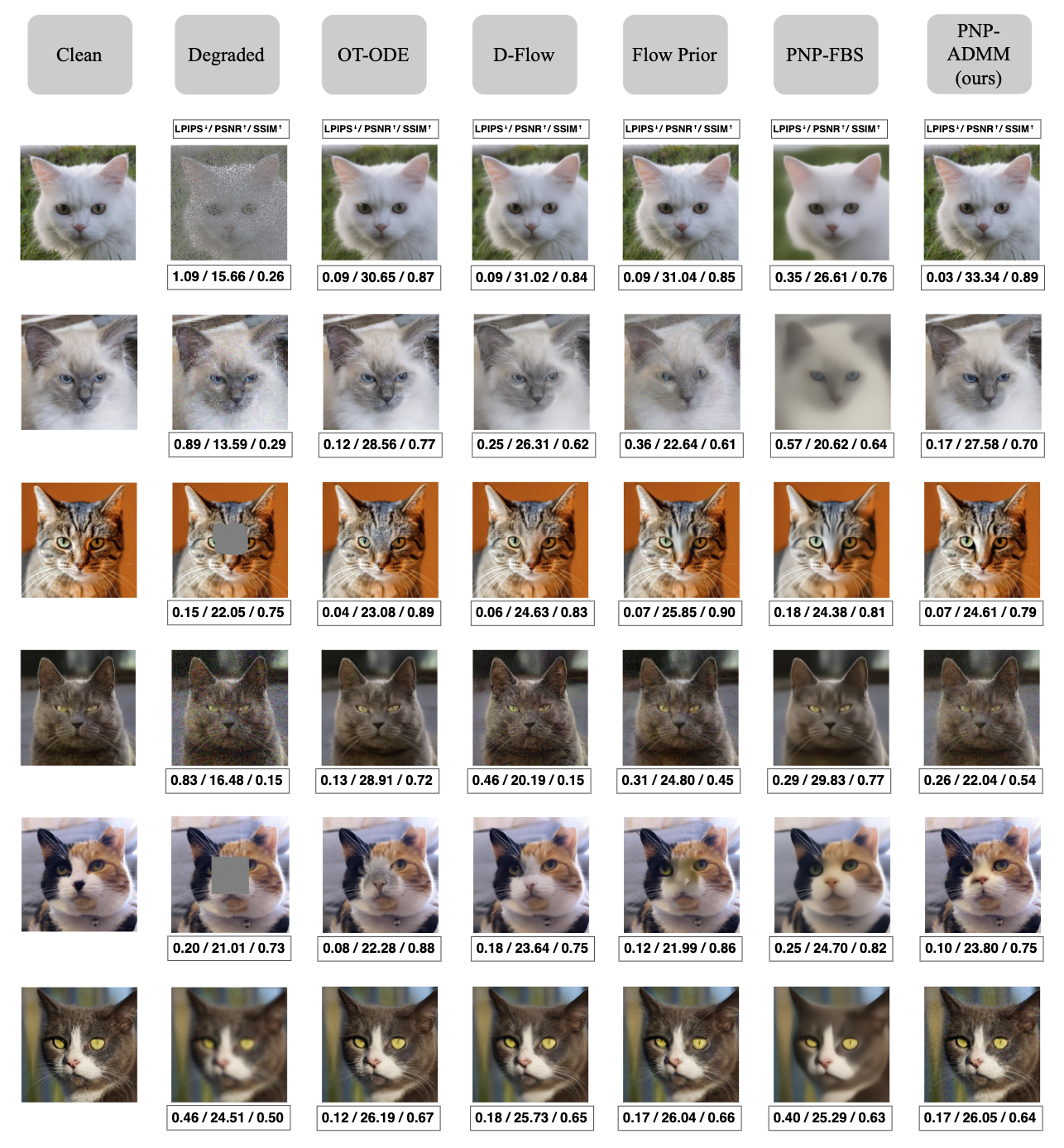}
  \caption{Additional qualitative comparisons on the AFHQ-Cat dataset across multiple inverse tasks, including 
  random inpainting (1st row), super-resolution (2nd row), paintbrush inpainting (3rd row), denoising (4th row), 
  box inpainting (5th row), and deblurring (6th row). 
From left to right: clean image, degraded input, OT-ODE, D-Flow, Flow Prior, PnP-FBS, and the proposed PnP-ADMM (ours). 
Quantitative metrics (LPIPS / PSNR/ SSIM) are reported below each reconstruction.}
\end{figure}

\clearpage

\begin{figure}[H]
    \centering
    \includegraphics[width=\textwidth,height=\textheight,keepaspectratio]{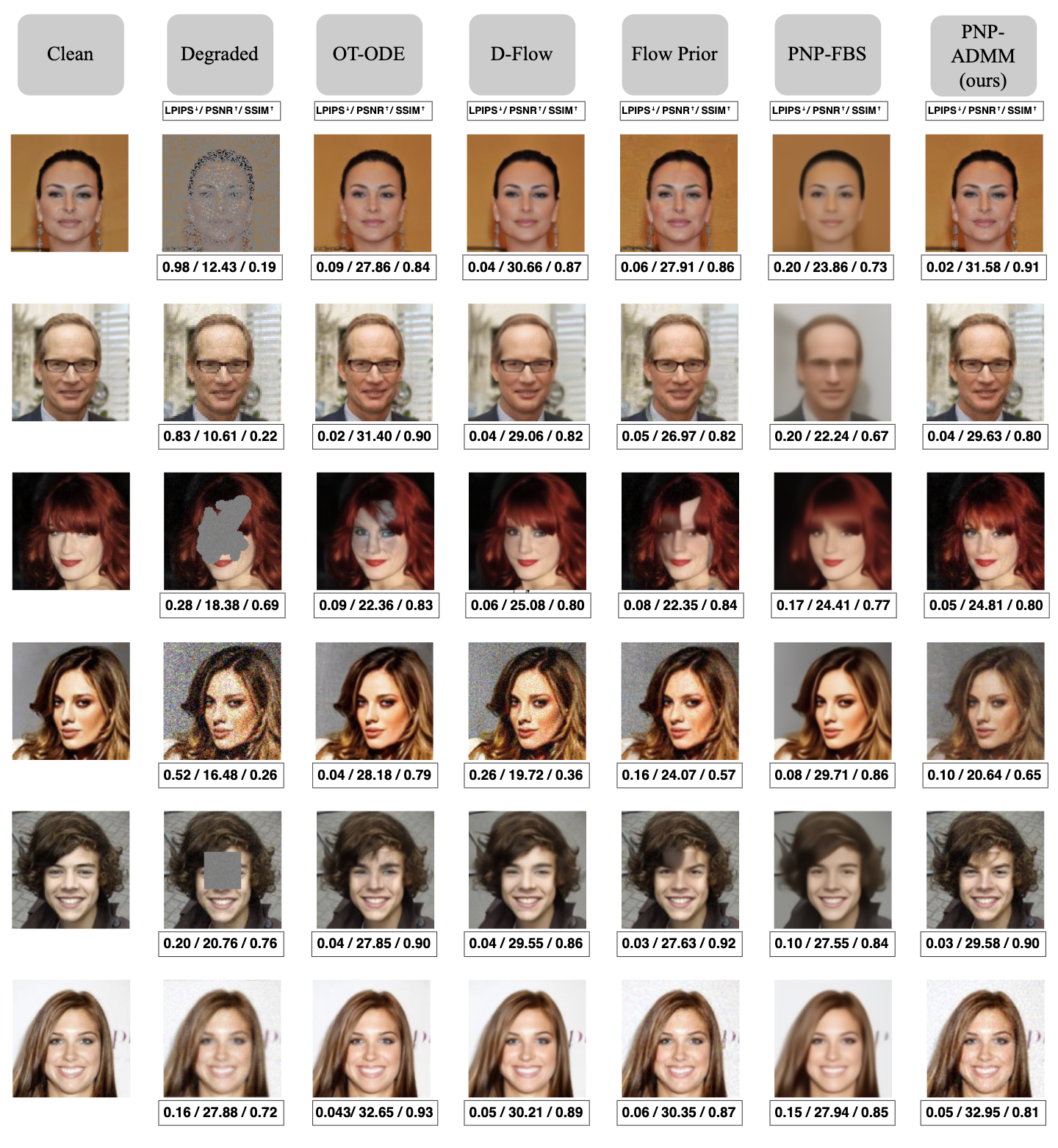}
    \caption{Additional qualitative comparisons on the CelebA dataset across multiple inverse tasks, including 
    random inpainting (1st row), super-resolution (2nd row), paintbrush inpainting (3rd row), denoising (4th row), 
  box inpainting (5th row), and deblurring (6th row). 
From left to right: clean image, degraded input, OT-ODE, D-Flow, Flow Prior, PnP-FBS, and the proposed PnP-ADMM (ours). 
Quantitative metrics (LPIPS / PSNR / SSIM) are reported below each reconstruction. 
    }
\end{figure}

\begin{figure}[H]
    \centering
    \includegraphics[width=\linewidth,height=0.9\textheight,keepaspectratio]{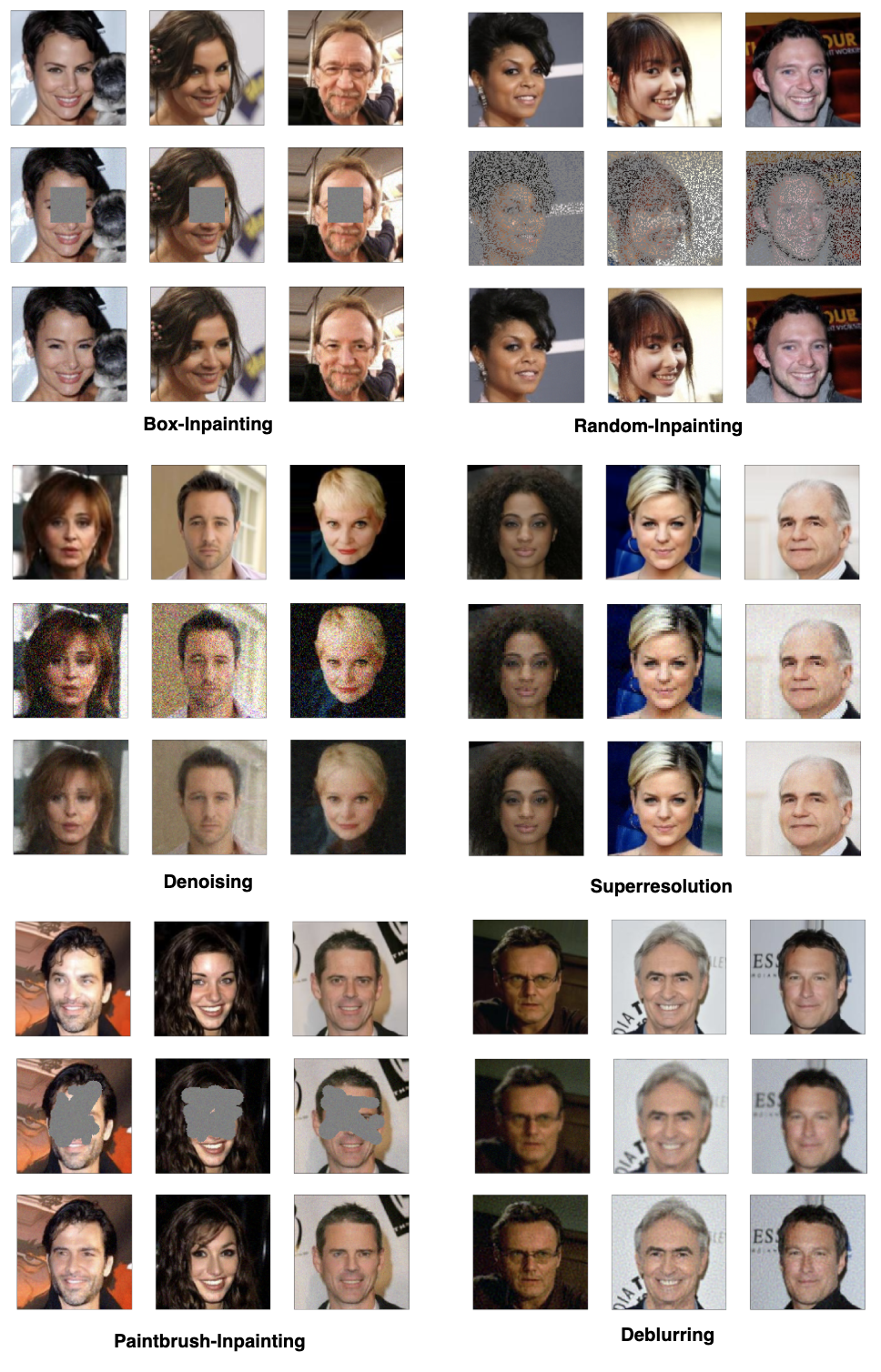}
    \caption{Supplementary results of our method on CelebA. For each inverse task: row 1 shows the clean image, row 2 the degraded input, and row 3 the reconstruction. }
\end{figure}

\begin{figure}[H]
    \centering
    \includegraphics[width=\linewidth,height=0.9\textheight,keepaspectratio]{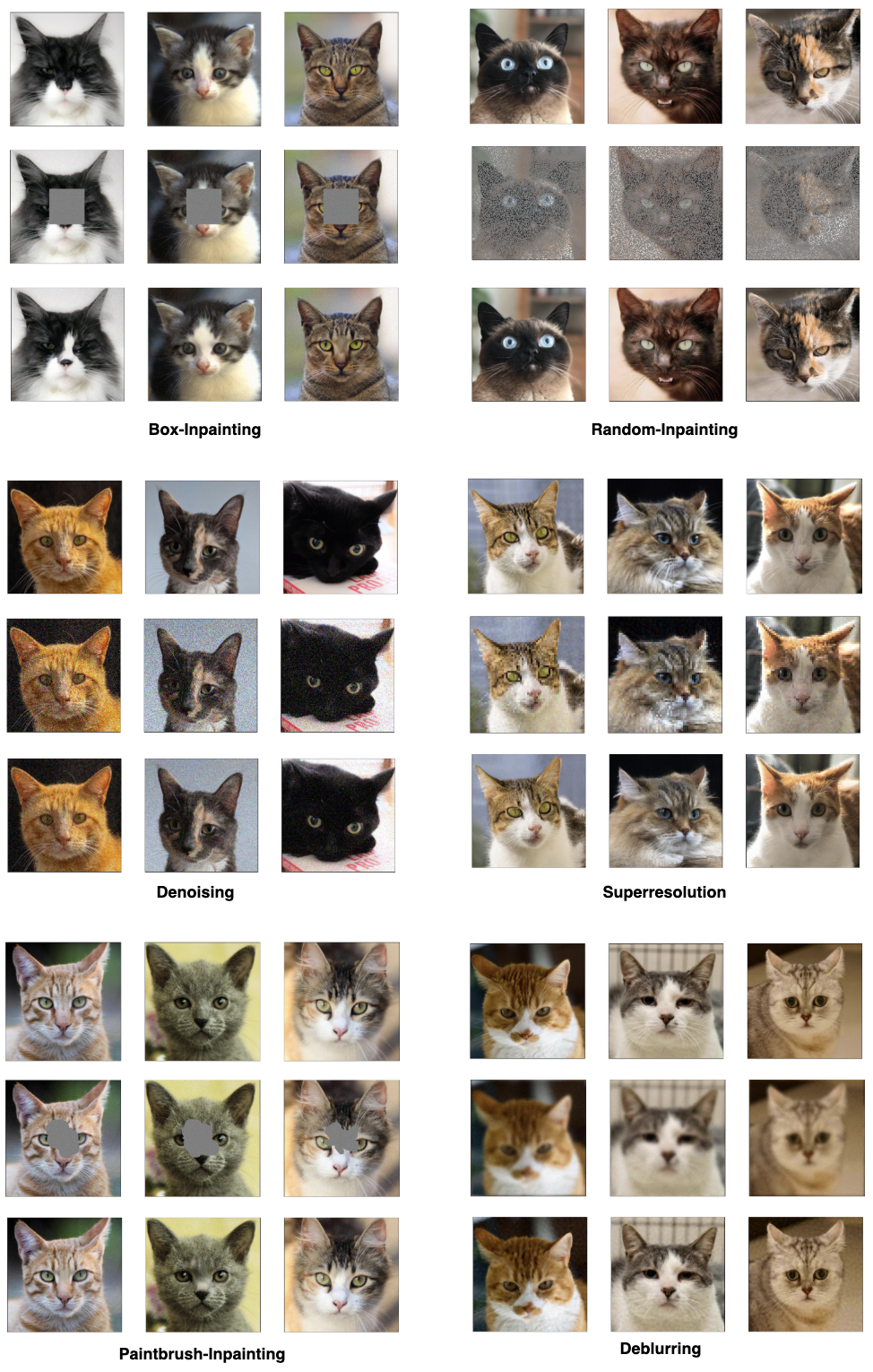}
    \caption{Supplementary results of our method on AFHQ-Cat. For each inverse task: row 1 shows the clean image, row 2 the degraded input, and row 3 the reconstruction.}
\end{figure}


\newpage
\section*{NeurIPS Paper Checklist}

\begin{enumerate}

\item {\bf Claims}
    \item[] Question: Do the main claims made in the abstract and introduction accurately reflect the paper's contributions and scope?
    \item[] Answer: \answerYes{} 
    \item[] Justification: The contributions are clearly mentioned in section \ref{intro} and the analysis is done in 
    sections \ref{analysisadmm} and \ref{computation}, as well as the results reported in 
    tables \ref{tab:results} and \ref{tab:time_memory_compact}.

\item {\bf Limitations}
    \item[] Question: Does the paper discuss the limitations of the work performed by the authors?
    \item[] Answer: \answerYes{} 
    \item[] Justification: We provide the limitations of this study in section \ref{conclusion} and in more 
    detail in Appendix \ref{theory_limits}.

\item {\bf Theory assumptions and proofs}
    \item[] Question: For each theoretical result, does the paper provide the full set of assumptions and a complete (and correct) proof?
    \item[] Answer: \answerYes{} 
    \item[] Justification: We discuss about the convergence of our method in section \ref{conv_analysis} 
    where we say that it is common to assume nonexpansiveness of denoiser in the PnP setting, therefore we 
    show the convergence empirically in appendix section \ref{app:conv}.

    \item {\bf Experimental result reproducibility}
    \item[] Question: Does the paper fully disclose all the information needed to reproduce the main experimental results of the paper to the extent that it affects the main claims and/or conclusions of the paper (regardless of whether the code and data are provided or not)?
    \item[] Answer: \answerYes{} 
    \item[] Justification: The paper provides detailed methodological and experimental descriptions in section \ref{experiments}, 
    and a code repository with a README is included, ensuring reproducibility of the results.

\item {\bf Open access to data and code}
    \item[] Question: Does the paper provide open access to the data and code, with sufficient instructions to faithfully reproduce the main experimental results, as described in supplemental material?
    \item[] Answer: \answerYes{} 
    \item[] Justification: The paper provides a code repository with open access, along with a README containing instructions to reproduce the main experimental results and access the required data.

\item {\bf Experimental setting/details}
    \item[] Question: Does the paper specify all the training and test details (e.g., data splits, hyperparameters, how they were chosen, type of optimizer) necessary to understand the results?
    \item[] Answer: \answerYes{} 
    \item[] Justification: We provide a comprehensive description of the experimental setup in Section \ref{experiments}. 
    In particular, this section clearly outlines the dataset used, the experimental settings, the baseline methods, 
    and the selected hyperparameters. 
    Details on the hyperparameter selection process are given in appendix section \ref{rho_analysis}.

\item {\bf Experiment statistical significance}
    \item[] Question: Does the paper report error bars suitably and correctly defined or other appropriate information about the statistical significance of the experiments?
    \item[] Answer: \answerNo{} 
    \item[] Justification: All results are averaged over multiple batches, and details are provided in Section \ref{experiments}. We do not include error bars as the evaluation involves computationally 
    expensive baselines, making multiple independent runs impractical; averaging over a large number of samples provides a stable estimate of performance.

\item {\bf Experiments compute resources}
    \item[] Question: For each experiment, does the paper provide sufficient information on the computer resources (type of compute workers, memory, time of execution) needed to reproduce the experiments?
    \item[] Answer: \answerYes{} 
    \item[] Justification: We provide the details on the GPU specifications in section \ref{computation} 
    and detailed time and memory consumption in table \ref{tab:time_memory_compact}.
    
\item {\bf Code of ethics}
    \item[] Question: Does the research conducted in the paper conform, in every respect, with the NeurIPS Code of Ethics \url{https://neurips.cc/public/EthicsGuidelines}?
    \item[] Answer: \answerYes{} 
    \item[] Justification: The research adheres to the NeurIPS Code of Ethics and standard ethical guidelines for scientific research and reporting.

\item {\bf Broader impacts}
    \item[] Question: Does the paper discuss both potential positive societal impacts and negative societal impacts of the work performed?
    \item[] Answer: \answerNA{} 
    \item[] Justification: This work is purely methodological and focuses on foundational research in solving inverse problems, without direct deployment or application to societal settings. 
    Therefore, it does not have immediate or identifiable societal impacts.
    
\item {\bf Safeguards}
    \item[] Question: Does the paper describe safeguards that have been put in place for responsible release of data or models that have a high risk for misuse (e.g., pre-trained language models, image generators, or scraped datasets)?
    \item[] Answer: \answerNA{} 
    \item[] Justification: This work does not release high-risk assets such as 
    large-scale generative models or sensitive datasets, and focuses on methodological 
    contributions. Therefore, no specific safeguards are required.

\item {\bf Licenses for existing assets}
    \item[] Question: Are the creators or original owners of assets (e.g., code, data, models), used in the paper, properly credited and are the license and terms of use explicitly mentioned and properly respected?
    \item[] Answer: \answerYes{} 
    \item[] Justification: The work builds upon an existing standard datasets/models and codebase, 
    and the original sources are properly cited. Their licenses and terms of use are respected in accordance 
    with the respective repositories and publications.

\item {\bf New assets}
    \item[] Question: Are new assets introduced in the paper well documented and is the documentation provided alongside the assets?
    \item[] Answer: \answerYes{} 
    \item[] Justification: A code repository is provided with the submission, along with sufficient 
    implementation details, enabling reproducibility and proper use of the proposed method.

\item {\bf Crowdsourcing and research with human subjects}
    \item[] Question: For crowdsourcing experiments and research with human subjects, does the paper include the full text of instructions given to participants and screenshots, if applicable, as well as details about compensation (if any)? 
    \item[] Answer: \answerNA{} 
    \item[] Justification: This work does not involve human subjects, user studies, or crowdsourced data, and is purely focused on algorithmic and theoretical development.

\item {\bf Institutional review board (IRB) approvals or equivalent for research with human subjects}
    \item[] Question: Does the paper describe potential risks incurred by study participants, whether such risks were disclosed to the subjects, and whether Institutional Review Board (IRB) approvals (or an equivalent approval/review based on the requirements of your country or institution) were obtained?
    \item[] Answer: \answerNA{} 
    \item[] Justification: This work does not involve human subjects, user studies, or crowdsourced data, and is purely focused on algorithmic and theoretical development.

\item {\bf Declaration of LLM usage}
    \item[] Question: Does the paper describe the usage of LLMs if it is an important, original, or non-standard component of the core methods in this research? Note that if the LLM is used only for writing, editing, or formatting purposes and does \emph{not} impact the core methodology, scientific rigor, or originality of the research, declaration is not required.
    \item[] Answer: \answerNA{} 
    \item[] Justification: The core method developed in this research does not involve use of LLMs.

\end{enumerate}

\end{document}